\documentclass{article} 
\usepackage{iclr2026_conference}
\usepackage{times}
\iclrfinalcopy

\usepackage{amsmath,amsfonts,bm}

\def\eqref#1{equation~\ref{#1}}

\def\1{\bm{1}}

\DeclareMathAlphabet{\mathsfit}{\encodingdefault}{\sfdefault}{m}{sl}
\SetMathAlphabet{\mathsfit}{bold}{\encodingdefault}{\sfdefault}{bx}{n}

\usepackage{hyperref}

\makeatletter
\newcommand\blfootnote[1]{%
  \begingroup
  \renewcommand\thefootnote{}\relax
  \Hy@raisedlink{\footnotemark}%
  \footnotetext{#1}%
  \addtocounter{footnote}{-1}%
  \endgroup
}
\makeatother

\usepackage{url}
\usepackage{booktabs}
\usepackage{multirow}
\usepackage{booktabs,multirow,tabularx,array}
\usepackage{makecell}
\usepackage{tikz}
\usepackage{float}
\newcolumntype{C}{>{\centering\arraybackslash}X}

\usepackage{booktabs,multirow,tabularx,array,makecell}

\usepackage{xcolor,booktabs}

\newcommand{\bdiff}[1]{\textcolor{blue}{\scriptsize #1}}
\newcommand{\rdiff}[1]{\textcolor{red}{\scriptsize #1}} 
\newcommand{\inlineDiff}[2]{#1\ #2}

\usepackage{array}
\newcolumntype{B}{>{\color{blue}}c}
\newcolumntype{R}{>{\color{red}}c}
\usepackage[most]{tcolorbox}
\tcbset{colback=green!10!white, colframe=green!50!black, boxrule=0.5pt, arc=3pt}
\usepackage{booktabs,multirow,makecell,xcolor}

\usepackage{tikz}
\usetikzlibrary{arrows.meta}
\tikzset{chatarr/.style={-{Stealth[length=2.4mm]}, semithick}}
\usepackage{tcolorbox}
\usepackage{amsmath}

\newcommand{\filledcircle}[1]{\tikz[baseline=(char.base)]{%
    \node[shape=circle,fill=black,text=white,inner sep=0.5pt,minimum size=3.5mm] (char) {\tiny\bfseries#1};}}

\newcommand{\methodname}[0]{\texttt{CORE}}
\usepackage{graphicx}

\title{\methodname{}: Concept-Oriented Reinforcement for Bridging the Definition–Application Gap in Mathematical Reasoning}

\author{
Zijun Gao$^{1}$,
Zhikun Xu$^{2}$,
Xiao Ye$^{2}$,
Ben Zhou$^{2}$\\
$^{1}$University of Illinois Urbana--Champaign \quad
$^{2}$Arizona State University
}

\begin{document}

\maketitle

\begingroup
\makeatletter
\renewcommand{\thefootnote}{}
\renewcommand{\theHfootnote}{\arabic{footnote}}
\footnotetext[0]{Correspondence to:
\href{mailto:zijung3@illinois.edu}{zijung3@illinois.edu} and
\href{mailto:zhikunxu@asu.edu}{zhikunxu@asu.edu}.}
\makeatother
\endgroup

\begin{abstract}
Large language models (LLMs) often solve challenging math exercises yet fail to apply the concept right when the problem requires genuine understanding. Popular Reinforcement Learning with Verifiable Rewards (RLVR) pipelines reinforce final answers but provide little fine-grained conceptual signal, so models improve at pattern reuse rather than conceptual applications. We introduce \methodname{} (Concept-Oriented REinforcement), an RL training framework that turns explicit concepts into a controllable supervision signal. Starting from a high-quality, low-contamination textbook resource that links verifiable exercises to concise concept descriptions, we run a sanity probe showing LLMs can restate definitions but fail concept-linked quizzes, quantifying the conceptual reasoning gap. \methodname{} then (i) synthesizes concept-aligned quizzes, (ii) injects brief concept snippets during rollouts to elicit concept-primed trajectories, and (iii) reinforces conceptual reasoning via trajectory replacement after group failures, a lightweight forward-KL constraint that aligns unguided with concept-primed policies, or standard GRPO directly on concept-aligned quizzes. Across several models, \methodname{} delivers consistent gains over vanilla and SFT baselines on both in-domain concept--exercise suites and diverse out-of-domain math benchmarks. \methodname{} unifies direct training on concept-aligned quizzes and concept-injected rollouts under outcome regularization. It provides fine-grained conceptual supervision that bridges problem-solving competence and genuine conceptual reasoning, while remaining algorithm- and verifier-agnostic. Code is available at \url{https://github.com/ARC-ASU/CORE}.
\end{abstract}

\section{Introduction}
Recent LLMs are becoming good at tackling competition-level questions, yet they fall short of conceptual math reasoning beyond applying competition tricks or executing numerical calculations \citep{yang2024qwen25mathtechnicalreportmathematical,guo2025mathematicalprooflitmustest,huang2025gemini25procapable,chen2025seedproverdeepbroadreasoning}. Here, conceptual reasoning means identifying the right concept and applying it in the solution, as opposed to procedural pattern matching often sufficient for GSM8K \citep{cobbe2021gsm8k} or MATH \citep{hendrycks2021measuring} and exposed by perturbation-based tests \citep{patel-etal-2021-nlp,yu2024reasonagainusingextractablesymbolic,mirzadeh2025gsmsymbolicunderstandinglimitationsmathematical,huang2025mathperturb}  On many benchmarks, models can mimic solution templates, chain together routine algebraic steps, and even memorize recurring patterns—while still choosing the wrong concept for a problem or failing to correctly apply certain concepts. This gap matters: solving a word problem by spotting a familiar cue is not the same as understanding linear independence, continuity, or convexity and deploying those notions correctly \citep{li2025one, huang2025mathperturb}.

Two main factors are contributing to this gap. First, large collections of exercise-style problems can often be solved by exploiting surface regularities (formats, keywards, and step patterns) rather than engaging the intended mathematical concepts \citep{guo2025rightenoughpitfallsoutcome,wu2025reasoningmemorizationunreliableresults}. Second, widely used RLVR pipelines \citep{schulman2017proximalpolicyoptimizationalgorithms,shao2024deepseekmathpushinglimitsmathematical} optimize a terminal scalar reward for correctness. Such signals can improve search heuristics \citep{qin2025decomposingelementsproblemsolving} but are too coarse to specify which concept should be invoked, where it should enter the argument, or how it supports subsequent steps. Moreover, directly assessing conceptual understanding is also challenging: concepts cannot be instantiated without a clear motivation or goal. We operationalize this by pairing concise concept definitions with aligned quiz questions that necessitate the concept. In a preliminary diagnostic, models readily recites the concepts yet frequently fail the quizzes, revealing a pronounced \textbf{definition-application gap}.

In order to mitigate this definition-application gap, we propose \methodname{} (Concept-Oriented REinforcement), an RL-based framework that turns explicit mathematical concepts into concept-driven training signals in sampling. \methodname{} starts from curating a high-quality data that (i) provide human-verified exercises and (ii) link each exercise to the underlying concept(s), which serve as the in-domain test set and seed data for further generation and training signals. We then expand coverage by generating additional concept-aligned quizzes using LLMs to curate the training set. For the training recipe, \methodname{} has explored three designs: \filledcircle{1} \textbf{Original RL (\methodname{}-Base):} directly training with the generated quizzes by RL algorithms, \filledcircle{2} \textbf{Concept Enhancement (\methodname{}-CR):} injecting concise concept snippets into rollouts to replace part of the original ones when all trajectories are incorrect, and \filledcircle{3} \textbf{KL Divergence (\methodname{}-KL):} implementing the KL divergence term between the concept-guided trajectories and the original ones to implicitly constraint the model towards using concepts. The above design choices investigate in three main components, \textbf{data}, \textbf{rollouts}, and \textbf{loss function}, of RL algorithms to mitigate the definition-application gap and improve the conceptual reasoning. Moreover, this framework wraps around standard policy-gradient reinforcement algorithms without architectural changes. At test time, the trained model is evaluated without providing the concept text, measuring whether concept-aware training translates into genuine conceptual competence rather than reasoning shortcuts.

Empirical results show that \methodname{} delivers consistent and significant improvements across \textbf{Qwen2\text{-}Math\text{-}7B}~\citep{yang2024qwen2technicalreport}, \textbf{DeepSeek\text{-}R1\text{-}Distill\text{-}Qwen\text{-}1.5B} \citep{deepseek2025r1}, \textbf{Qwen2.5\text{-}Math\text{-}1.5B}\citep{yang2024qwen2technicalreport} and \textbf{Llama\text{-}3\text{-}8B\text{-}Instruct}~\citep{grattafiori2024llama3herdmodels}. On Qwen2\text{-}Math\text{-}7B, CORE variants yield gains of up to \textbf{9.3\%} on \textsc{Textbook} and \textbf{9.6\%} on \textsc{TheoremQA}, highlighting improved conceptual alignment and reasoning depth. 
For DeepSeek\text{-}R1\text{-}DQ\text{-}1.5B, \textsc{CORE-CR} achieves \textbf{1.3\%} on \textsc{MMLU-STEM} and \textbf{1.2\%} on \textsc{SVAMP}, indicating stronger stability under out-of-domain reasoning. 
On Qwen2.5\text{-}Math\text{-}1.5B, CORE enhances \textsc{Minerva Math} by \textbf{3.3\%} and \textsc{TabMWP} by \textbf{1.9\%}, demonstrating its adaptability to specialized math domains. 
Finally, for Llama\text{-}3\text{-}8B\text{-}Instruct, \textsc{CORE-CR} surpasses Vanilla by up to \textbf{3.3\%} on \textsc{TabMWP} and also yields consistent improvements on \textsc{MMLU-STEM} and \textsc{SVAMP}. These results demonstrate that, without any architectural modifications, \methodname{} consistently enhances conceptual understanding and reasoning ability through explicit concept injection and concept-aware optimization.


\section{Related Works}
\paragraph{Mathematical Reasoning in LLMs}

Recent systems approach math reasoning through specialized training or sheer scale. WizardMath \citep{luo2025wizardmathempoweringmathematicalreasoning} uses reinforcement learning from Evol-Instruct feedback. MAmmoTH \citep{yue2023mammothbuildingmathgeneralist} blends chain-of-thought and program-of-thought for hybrid tuning. Qwen2.5-Math \citep{yang2024qwen25mathtechnicalreportmathematical} continues from a general model on a large curated math corpus. Llemma \citep{azerbayev2024llemmaopenlanguagemodel} is pre-trained on Proof-Pile. DeepSeekMath \citep{shao2024deepseekmathpushinglimitsmathematical} adds about 120B math tokens and introduces \textit{Group Relative Policy Optimization} (GRPO). InternLM2-Math \citep{ying2024internlmmath} unifies chain-of-thought, reward modeling, and formal reasoning. General-purpose models also lean heavily into math: Llama 3.1 \citep{grattafiori2024llama3herdmodels} up-samples math data, DeepSeek-R1 and DeepSeek-V3 \citep{deepseekai2025deepseekr1incentivizingreasoningcapability,deepseekai2025deepseekv3technicalreport} raise math and programming proportions across trillions of tokens, Claude 3.7 \citep{anthropic2025claude37} emphasizes transparent multi-step reasoning, Gemini \citep{comanici2025gemini25pushingfrontier} targets long-chain deduction, and OpenAI o1 and o3 \citep{openai2024openaio1card,openai2025o3} scale test-time compute. Despite these advances, many pipelines reward final answers or rely on data scale, leaving concept selection and application under-taught. Our \methodname{} addresses this by injecting explicit concept signals into rollouts and regularizing outcomes, yielding consistent gains on concept-dependent evaluations while remaining agnostic to the underlying RL algorithm.

\paragraph{Mathematical Benchmarks}
A wide range of math benchmarks now probes reasoning at various levels. At the elementary level, GSM8k is the standard for multi-step math problems. MAWPS \citep{koncel-kedziorski-etal-2016-mawps} aggregates sources, ASDiv \citep{miao2021diversecorpusevaluatingdeveloping} adds type and grade labels, and SVAMP \citep{patel-etal-2021-nlp} stresses robustness through controlled perturbations. Cross-lingual coverage includes CMATH for Chinese primary school \citep{wei2023cmathlanguagemodelpass} and CN Middle School 24, while standardized suites such as Gaokao 2023 EN, SAT Math, and MMLU-STEM enable broader STEM-wide comparisons \citep{zhong2023agieval,hendrycks2021measuring}. For competition-level reasoning, MATH curates Olympiad and contest problems with stepwise solutions, and OlympiadBench \citep{he-etal-2024-olympiadbench} extends to bilingual and multimodal settings that emphasize proof-style reasoning and reduce contamination risk. Our evaluations of \methodname{} are based on in-domain concept–exercise suites and out-of-domain math benchmarks, showing consistent gains over strong baselines and highlighting improvements specifically on concept-dependent categories.

\paragraph{Conceptual Reasoning}
Answer accuracy alone doesn’t reveal whether the right concepts were selected and used, several works probe whether models actually select and use the right concepts. Specifically in math, conceptual reasoning requires people to reason around math concepts and axioms at the play of math hypothesis, statements and problems \citep{simon2011studying}. \textsc{TheoremQA} \citep{chen-etal-2023-theoremqa} targets theorem application across STEM, explicitly requiring mapping from a named theorem to its correct use in problem solving. \textsc{GSM-Symbolic} \citep{mirzadeh2025gsmsymbolicunderstandinglimitationsmathematical} examines symbolic generalization limits of models trained on GSM8K-style data. \textsc{CounterMATH} \citep{li2025one} proposes counterexample-driven, concept-sensitive evaluations to diagnose superficial cues versus true concept use. Complementary directions include a conceptualization framework that maps abstract questions into verifiable symbolic programs \citep{zhou2024conceptualunbiasedreasoninglanguage}, a self-supervised analogical learning scheme that transfers high-level solutions across cases \citep{zhou2025selfsupervisedanalogicallearningusing}, and a Bayesian inference formulation coupling abductive proposals with structured deduction for calibrated decisions \citep{feng2025birdtrustworthybayesianinference}. 
Overall, these efforts indicate that a single end-point score is insufficient for assessing whether models select and correctly apply concepts; concept-aligned training signals or structured evaluation protocols are required.

\section{\methodname{}: Concept-Oriented Reinforcement Learning}
\label{tag:method}
\subsection{Overview}
We study math reasoning where success depends mainly on conceptual math reasoning rather than replaying surface templates in the training data. Our proposed framework \methodname{} has been developed through the following stages: \textbf{dataset curation}, \textbf{gap diagnostics}, and \textbf{concept reinforcement recipe}. For \textit{dataset curation}, we have leveraged a classical mathematical textbook with clear associations between concepts and exercises for training and evaluation. For \textit{gap diagnostics}, we have used the curated data to both qualitatively and quantitatively identify the definition-application gaps in the conceptual mathematical reasoning. For \textit{concept reinforcement recipe}, we have mainly designed three training recipes in reinforcing the models' conceptual mathematical reasoning.

\subsection{Dataset Curation}
To acquire rigorous conceptual reasoning signals, we curated a corpus from a canonical textbook, 
\textit{Advanced Algebra} (3rd Edition) \citep{fudanalgebra}. This source was chosen for two-fold reasons. First, it provides 
a comprehensive and structured curriculum in linear algebra, progressing logically from foundational 
concepts like determinants and matrices to advanced topics such as linear spaces and canonical forms. 
Its ten chapters are methodically structured, each containing: 
i) core concept definitions $(\mathcal{C})$, 
ii) illustrative examples, and 
iii) concept-aligned exercises $(\mathcal{E})$, 
where exercises in a given chapter primarily test the concepts introduced in that same chapter. The textbook's long-standing use and human verification ensure a logical progression of topics and coherent conceptual dependencies, making it an ideal corpus for developing and evaluating a concept-oriented learning paradigm. Second, by manually translating this Chinese textbook into English, we significantly mitigate the risk of training data contamination present in many existing English-language corpora.

The extraction yielded 236 concept texts, 703 examples, and 140 multiple-choice questions 
sourced from the exercise sections. More details are provided in Appendix~\ref{appendix:data}.

\subsection{Gap Diagnostics}
\subsubsection{Probing the Gap Between Knowledge Recitation and Application}
The structured nature of our curated corpus, with its explicit mapping between concepts $(\mathcal{C})$ 
and exercises $(\mathcal{E})$, provides an ideal testbed to diagnose a critical failure mode in 
state-of-the-art LLMs: the disconnect between parametric knowledge and its application in problem-solving. 
We conduct the following sanity check to probe this gap, which provides the foundational motivation for 
our proposed training framework.

\begin{figure}[t]
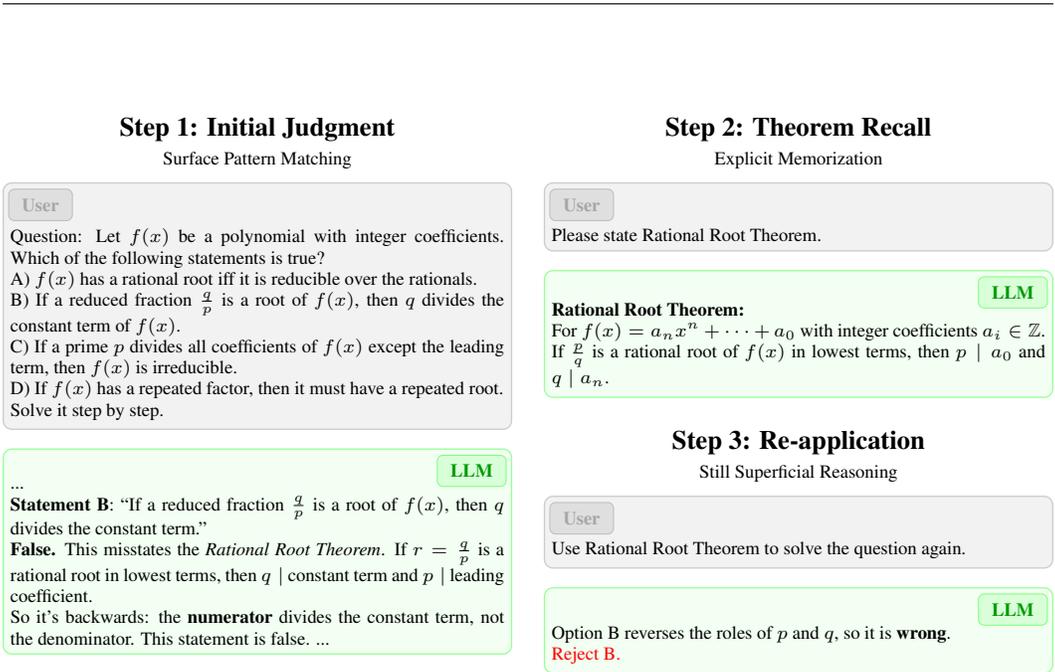

\centering

\begin{minipage}[t]{0.485\columnwidth}
\centering
\vspace{0pt}

\textbf{Step 1: Initial Judgment}\\
{\scriptsize Surface Pattern Matching}

\vspace{-2pt}

\begin{tcolorbox}[
enhanced,
colback=gray!10,
colframe=gray!40,
fontupper=\scriptsize,
boxsep=0.5pt,
left=2pt,right=2pt,
top=5.8mm,bottom=1mm,
overlay={
\node[anchor=north west,
      xshift=-0.5mm,
      yshift=0.4mm]
at (frame.north west)
{\tcbox[
  on line,
  colback=gray!25,
  colframe=gray!50,
  boxrule=0.4pt,
  arc=2pt,
  left=2pt,right=2pt,top=0.5pt,bottom=0.5pt
]{\scriptsize\bfseries\textcolor{gray!70}{User}}};
}
]
Question: Let $f(x)$ be a polynomial with integer coefficients. Which of the following statements is true?

A) $f(x)$ has a rational root iff it is reducible over the rationals.

B) If a reduced fraction $\tfrac{q}{p}$ is a root of $f(x)$, then $q$ divides the constant term of $f(x)$.

C) If a prime $p$ divides all coefficients of $f(x)$ except the leading term, then $f(x)$ is irreducible.

D) If $f(x)$ has a repeated factor, then it must have a repeated root.

Solve it step by step.
\end{tcolorbox}

\vspace{-4pt}

\begin{tcolorbox}[
enhanced,
colback=green!5,
colframe=green!40,
fontupper=\scriptsize,
boxsep=0.5pt,
left=2pt,right=2pt,
top=4.8mm,bottom=1mm,
overlay={
\node[anchor=north east,
      xshift=0.5mm,
      yshift=0.4mm]
at (frame.north east)
{\tcbox[
  on line,
  colback=green!15,
  colframe=green!50,
  boxrule=0.4pt,
  arc=2pt,
  left=2pt,right=2pt,top=0.5pt,bottom=0.5pt
]{\scriptsize\bfseries\textcolor{green!60!black}{LLM}}};
}
]
...

\textbf{Statement B}: ``If a reduced fraction $\tfrac{q}{p}$ is a root of $f(x)$, then $q$ divides the constant term.''

\textbf{False.} This misstates the \textit{Rational Root Theorem}. If $r=\tfrac{q}{p}$ is a rational root in lowest terms, then $q \mid$ constant term and $p \mid$ leading coefficient.

So it’s backwards: the \textbf{numerator} divides the constant term, not the denominator. This statement is false.
...
\end{tcolorbox}

\end{minipage}
\hfill
\begin{minipage}[t]{0.485\columnwidth}
\centering
\vspace{0pt}

\textbf{Step 2: Theorem Recall}\\
{\scriptsize Explicit Memorization}

\vspace{-2pt}

\begin{tcolorbox}[
enhanced,
colback=gray!10,
colframe=gray!40,
fontupper=\scriptsize,
boxsep=0.5pt,
left=2pt,right=2pt,
top=5.8mm,bottom=1mm,
overlay={
\node[anchor=north west,
      xshift=-0.5mm,
      yshift=0.4mm]
at (frame.north west)
{\tcbox[
  on line,
  colback=gray!25,
  colframe=gray!50,
  boxrule=0.4pt,
  arc=2pt,
  left=2pt,right=2pt,top=0.5pt,bottom=0.5pt
]{\scriptsize\bfseries\textcolor{gray!70}{User}}};
}
]
Please state Rational Root Theorem.
\end{tcolorbox}

\vspace{-4pt}

\begin{tcolorbox}[
enhanced,
colback=green!5,
colframe=green!40,
fontupper=\scriptsize,
boxsep=0.5pt,
left=2pt,right=2pt,
top=4.0mm,bottom=1mm,
overlay={
\node[anchor=north east,
      xshift=0.5mm,
      yshift=0.4mm]
at (frame.north east)
{\tcbox[
  on line,
  colback=green!15,
  colframe=green!50,
  boxrule=0.4pt,
  arc=2pt,
  left=2pt,right=2pt,top=0.5pt,bottom=0.5pt
]{\scriptsize\bfseries\textcolor{green!60!black}{LLM}}};
}
]
\textbf{Rational Root Theorem:}

For $f(x)=a_nx^n+\dots+a_0$ with integer coefficients $a_i\in\mathbb{Z}$.

If $\tfrac{p}{q}$ is a rational root of $f(x)$ in lowest terms, then $p\mid a_0$ and $q\mid a_n$.
\end{tcolorbox}

\vspace{6pt}

\textbf{Step 3: Re-application}\\
{\scriptsize Still Superficial Reasoning}

\vspace{-2pt}

\begin{tcolorbox}[
enhanced,
colback=gray!10,
colframe=gray!40,
fontupper=\scriptsize,
boxsep=0.5pt,
left=2pt,right=2pt,
top=5.8mm,bottom=1mm,
overlay={
\node[anchor=north west,
      xshift=-0.5mm,
      yshift=0.4mm]
at (frame.north west)
{\tcbox[
  on line,
  colback=gray!25,
  colframe=gray!50,
  boxrule=0.4pt,
  arc=2pt,
  left=2pt,right=2pt,top=0.5pt,bottom=0.5pt
]{\scriptsize\bfseries\textcolor{gray!70}{User}}};
}
]
Use Rational Root Theorem to solve the question again.
\end{tcolorbox}

\vspace{-4pt}

\begin{tcolorbox}[
enhanced,
colback=green!5,
colframe=green!40,
fontupper=\scriptsize,
boxsep=0.5pt,
left=2pt,right=2pt,
top=4.8mm,bottom=1mm,
overlay={
\node[anchor=north east,
      xshift=0.5mm,
      yshift=0.4mm]
at (frame.north east)
{\tcbox[
  on line,
  colback=green!15,
  colframe=green!50,
  boxrule=0.4pt,
  arc=2pt,
  left=2pt,right=2pt,top=0.5pt,bottom=0.5pt
]{\scriptsize\bfseries\textcolor{green!60!black}{LLM}}};
}
]
Option B reverses the roles of $p$ and $q$, so it is \textbf{wrong}.

\textcolor{red}{Reject B.}
\end{tcolorbox}

\end{minipage}

\vspace{-4pt}

\caption{An example of ChatGPT-4o’s superficial understanding of the Rational Root Theorem. Although the model correctly recalls the theorem, it fails to engage in structural verification across reasoning steps.}
\label{fig:example}
\end{figure}

We perform a qualitative analysis on the multiple-choice exercises that a powerful baseline, GPT-4o\footnote{\url{https://openai.com/index/hello-gpt-4o/}}, 
failed due to conceptual errors. After a model fails an exercise, we prompt it to describe 
    the core mathematical concept associated with that problem (e.g., ``Describe the concept of Linear 
    Independence.''). We observe that in the majority of cases, the model can accurately and comprehensively 
    recite the correct definition and properties of the concept, indicating that the requisite knowledge is 
    parametrically encoded. A classic example is shown in Figure~\ref{fig:example}. GPT-4o is able to correctly restate the Rational Root Theorem. However, when the numerator and denominator in the problem are swapped, the model still follows its ingrained reasoning and makes an incorrect judgment.

These sanity checks reveals a critical failure mode: the model's problem-solving process appears to be locked 
into a rigid, pattern-matching heuristic, failing to flexibly incorporate or be guided by explicit 
conceptual knowledge, even when it is readily available. This observed gap between the ability to recite 
a concept and the ability to apply it motivates the need for a training paradigm that explicitly 
forces the model to ground its reasoning in concepts, which we introduce in the following section.

\subsubsection{Synthetic Concept Probes and Robustness Evaluation}
\paragraph{Measuring Conceptual Understanding via Concept Probes.}  
A fundamental challenge in evaluating mathematical reasoning is the difficulty of quantitatively 
measuring a model's grasp of specific concepts. High-level benchmark scores often obscure 
fine-grained conceptual failures. The highly structured nature of our curated textbook, however, 
provides a unique opportunity to address this challenge. To create a direct and quantifiable 
measure of conceptual understanding, we introduce the idea of \textbf{Concept Probes}: targeted 
quizzes generated directly from the textbook's concept definitions and theorems, where a model's performance 
on these probes serves as a proxy for its mastery of the underlying concepts.

To realize this, we constructed a new dataset of conceptual quizzes. This process involved two 
key stages: generation and validation.

\begin{enumerate}
    \item \textbf{Generation:} We prompted a powerful generator model, Qwen2.5-72B-Instruct, 
    to create 5--8 multiple-choice quizzes for each of the 236 concept texts in our corpus. 
    This resulted in a candidate pool of 1,200 quizzes, each designed to be 
    closely tied to its source concept and formatted with standard \LaTeX.

    \item \textbf{Validation:} To ensure the quality and validity of these synthetic quizzes, 
    we designed a rigorous filtering pipeline using a separate, powerful assessor model, GPT-4o. 
    This cross-model validation strategy is intentionally designed to reduce harvester bias. 
    For each quiz, GPT-4o evaluated six dimensions (e.g., clarity, correctness, uniqueness) and 
    provided an overall rating and a confidence level. We discarded the 90 quizzes that were 
    rated ``Fair'' or ``Poor'' with high confidence. This stringent process yielded a final set 
    of \textbf{1,110 high-quality quizzes} that serve as our Concept Probes.
\end{enumerate}

\paragraph{Diagnostic Experiment: Robustness to Superficial Perturbations.}  
To validate our hypothesis that models rely on superficial heuristics, we designed a diagnostic 
experiment to test the robustness of their conceptual knowledge. Using our 1,110 curated quizzes, 
we use a \textbf{Robust Evaluation} protocol. For each quiz, we generate three variants by 
randomly permuting the order of its multiple-choice options. A model is considered to have 
\textit{robustly} solved a problem \textbf{only if} it correctly answers the original question 
\textit{and} all three of its permuted variants. This protocol is designed to test whether a model's 
understanding is invariant to semantically-irrelevant changes that preserve the core concept.

We applied both the standard and our Robust Evaluation protocols to a suite of 
contemporary models, including Qwen2-Math-7B, OLMo-7B-Instruct, and Llama-3-8B. The results, 
illustrated in Table~\ref{tab:main_results}, reveal a stark and consistent performance gap across all models. 
For instance, while a model like OLMo-2-7B may achieve high accuracy (e.g., $>70\%$) under 
the standard protocol, its performance plummets to below $50\%$ under Robust Evaluation. This 
significant degradation provides strong empirical evidence for our hypothesis, demonstrating that 
the models' success is heavily reliant on shallow heuristics rather than a 
deep, structural understanding of the underlying concepts.\\
\begin{table}[h!]
\centering
\caption{Performance comparison under Original vs. Robust Evaluation protocol across different models. The best performance in each column is highlighted in bold.}
\label{tab:main_results}
\begin{tabular}{l cc cc}
\toprule
& \multicolumn{2}{c}{\textbf{Standard evaluation}}
& \multicolumn{2}{c}{\textbf{Robust evaluation}} \\
\cmidrule(lr){2-3} \cmidrule(lr){4-5}
\textbf{Model} & \textbf{pass@1 accuracy} & \textbf{self-consistent} &
\textbf{pass@1 accuracy} & \textbf{self-consistent} \\
\midrule
Qwen-2-Math-7B &
\inlineDiff{74.33\%}{\bdiff{+0}} &
\inlineDiff{87.0\%}{\rdiff{+0}} &
\inlineDiff{45.92\%}{\bdiff{-28.4}} &
\inlineDiff{76.0\%}{\rdiff{-11.0}} \\
OLMo-2-7B      &
\inlineDiff{57.83\%}{\bdiff{+0}} &
\inlineDiff{70.17\%}{\rdiff{+0}} &
\inlineDiff{36.25\%}{\bdiff{-21.6}} &
\inlineDiff{44.42\%}{\rdiff{-25.8}} \\
LLaMA-3-8B     &
\inlineDiff{44.75\%}{\bdiff{+0}} &
\inlineDiff{70.92\%}{\rdiff{+0}} &
\inlineDiff{20.25\%}{\bdiff{-24.5}} &
\inlineDiff{46.75\%}{\rdiff{-24.2}} \\
\bottomrule
\end{tabular}
\end{table}

\subsection{Concept Reinforcement Recipe}
\begin{figure}[t]
    \centering
    \includegraphics[width=0.95\linewidth]{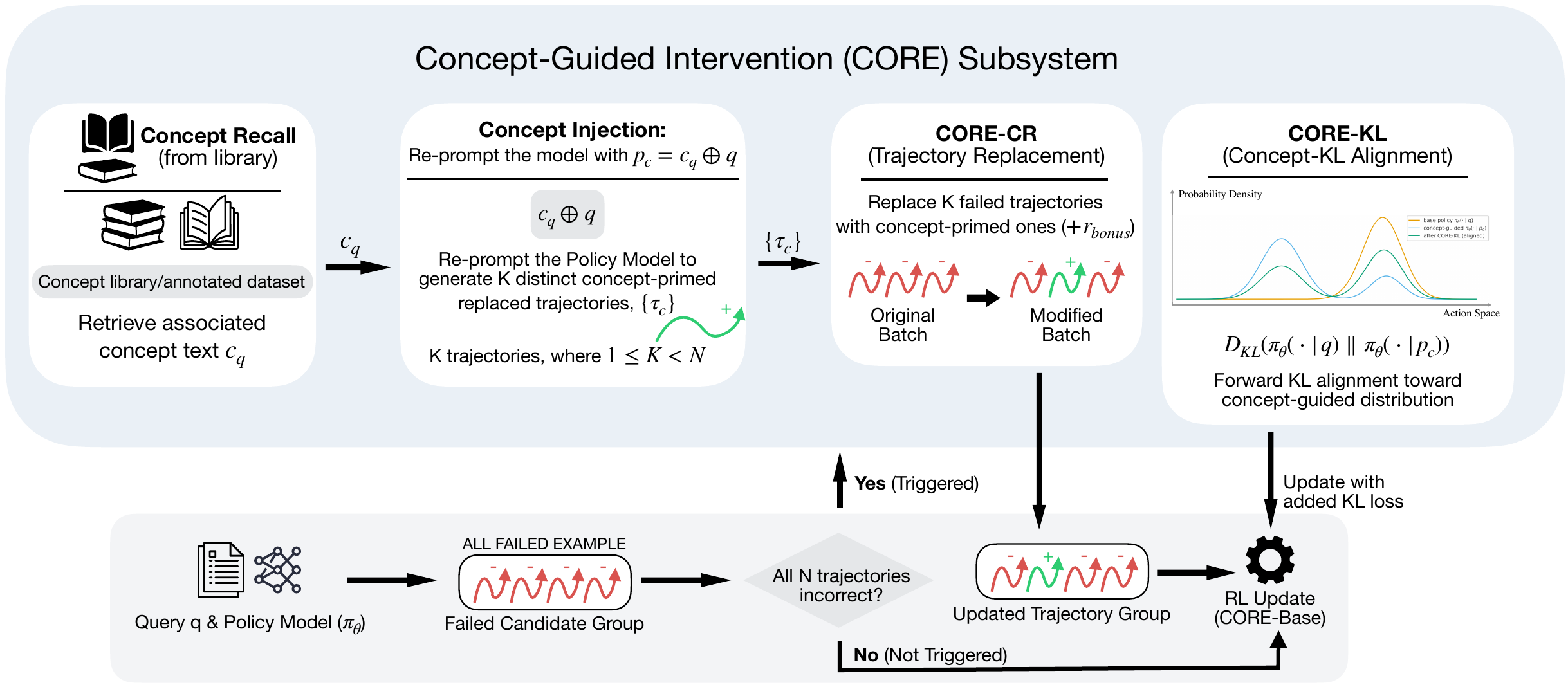}
    \caption{
    Overview of the Concept-Guided Reinforcement (\methodname{}) framework.
    For a given query, the policy model generates multiple candidate solutions. If any solution is correct, \methodname{}-Base proceeds.
    When all solutions fail, \methodname{} activates concept-guided correction:
    the Concept Recall module retrieves relevant domain knowledge, and Concept Injection
    re-prompts the model with this guidance to form corrected trajectories. \methodname{}-CR replaces failed paths with these concept-grounded ones to recover the learning signal, while \methodname{}-KL distills the concept-enhanced trajectories into the base policy via a forward KL loss.
    }
    \label{fig:teasure_figure}
\end{figure}

To bridge the gap between procedural mimicry and conceptual understanding, we propose \methodname{}, illustrated in Figure~\ref{fig:teasure_figure}. \methodname{} is an RL-based framework designed to inject conceptual knowledge into models through any policy gradient based RL algorithm. The core idea is to conditionally intervene during training with concept-guided instruction precisely when the model demonstrates a failure in understanding, guiding the policy update towards a more robust, concept-grounded reasoning process. Our \methodname{} framework mainly consists of the following three design choices, instantiated with the standard GRPO in this paper. Importantly, \methodname{} does not propose a new RL algorithm; GRPO is used here simply as a standard backbone. For completeness, we also discuss and evaluate a PPO-based variant in Appendix~\ref{sec:appendix-c1}.

\paragraph{\methodname{}-Base: Standard RL on Conceptual Quizzes.}The foundational approach within our framework, \textbf{\methodname{}-Base}, involves training the policy
$\pi_{\theta}$ directly on our curated set of conceptual quizzes $(\mathcal{Q})$ using the standard GRPO
algorithm. In this setting, the model learns from the conceptual data without any further explicit
guidance during the training process. This approach measures the model's ability to implicitly learn concepts
from the rich question-answer pairs generated from concepts.

\paragraph{\methodname{}-CR: {\itshape C}oncept-Guided Trajectory {\itshape R}eplacement.} Building upon the base setting, \textbf{\methodname{}-CR} introduces a conditional intervention triggered
by a \emph{conceptual failure event} (i.e., all $N$ responses in a GRPO group are incorrect).
Upon triggering, we form a concept-guided prompt
$p_c = c_q \oplus q$ where $q$ is the original problem from our quiz dataset, $c_q$ is its associated ground-truth concept text, and $\oplus$ denotes concatenation. We then generate $K$ new trajectories
$\{\tau_{c,1}, \ldots, \tau_{c,K}\}$ from the concept-guided policy, where $1 \leq K < N$.
We then \textbf{randomly select and replace} $K$ trajectories from the original failed group with these
new concept-guided ones.

To incentivize learning with concepts, we assign the new trajectories an augmented reward:
\[
R'(\tau_{c,j}) = R(\tau_{c,j}) + r_{\text{bonus}}
\]

where $r_{\text{bonus}} > 0$ is a hyperparameter. The GRPO update is then performed on this
partially replaced, concept-guided batch. Notably, there is a recent work called BREAD \citep{zhang2025breadbranchedrolloutsexpert}, which shares a very similar methodology to \methodname{}-CR while derives from rethinking the advantages of SFT and RL instead of improving conceptual reasoning. However, as empirically demonstrated in Section~\ref{sec:intrinsic_vs_distill}, \methodname{} achieves significant gains without distilling reasoning signals from a larger teacher model. This autonomy distinguishes \methodname{} from mainstream ``expert-anchored'' approaches like BREAD, which typically require superior external models to provide trajectory guidance.

\paragraph{\methodname{}-KL: Concept-Guided {\itshape KL}-Regularization.}   The \textbf{\methodname{}-KL} method introduces a fine-grained regularization signal to guide the policy's
internal reasoning process. This approach is also triggered by a \emph{conceptual failure event}. Instead of directly replacing trajectories, this method encourages the model's
standard step-by-step predictive process at each timestep $t$, denoted 
$\pi_{\theta}(\cdot \mid q, y_{<t})$, to align with the more robust process it exhibits 
when primed with a concept, $\pi_{\theta}(\cdot \mid p_c, y_{<t})$.

We formulate this as a \textbf{forward KL-divergence} objective. This choice is deliberate: it 
encourages the base policy to cover the full distribution of reasoning paths considered by the 
concept-guided ``teacher'' policy, rather than collapsing to a single mode, fostering a more 
comprehensive distillation of the entire reasoning process. Upon a conceptual failure trigger, we first sample a high-quality reference 
trajectory, $Y^* = (y^*_1, \ldots, y^*_T)$, from the \textit{current, online concept-guided policy}, 
i.e., $Y^* \sim \pi_\theta(\cdot \mid p_c)$. Our objective is then to minimize the KL-divergence 
between the next-token predictive distributions of the guided and un-guided policies at each timestep $t$, 
conditioned on the prefix of the reference trajectory $Y^*_{<t}$. This is formulated as a loss term 
added to the base RL objective:

\begin{equation}
\mathcal{L}_{\text{total}}
= \mathcal{L}_{\text{GRPO}} 
+ \lambda_{\text{KL}} \cdot \mathbb{E}_{Y^* \sim \pi_\theta(\cdot \mid p_c)} 
\left[
   \sum_{t=1}^{|Y^*|}
   D_{\text{KL}}\!\Big( \pi_\theta(\cdot \mid p_c, y^*_{<t}) \,\big\|\, \pi_\theta(\cdot \mid q, y^*_{<t}) \Big)
\right]
\end{equation}

where $\pi_\theta(\cdot \mid \text{context}, y^*_{<t})$ is the current policy's probability 
distribution over the next token. This forces the model's internal reasoning process on the 
original problem $q$ to faithfully mimic the process it would follow if it were explicitly given 
the concept $c_q$.

\methodname{}-Base primarily functions as a consolidation mechanism, reinforcing the application of concepts already encountered during pre-training, whereas \methodname{}-CR and \methodname{}-KL serve as complementary corrective strategies for concepts that are not yet reliably mastered, operating in parallel through explicit and implicit forms of concept intervention. Together, these variants constitute parallel and complementary ways of incorporating conceptual signals into reinforcement learning within a unified framework.



\section{Experiments}
\label{sec:experiments}
\subsection{Baseline Models}

We select \textbf{Qwen2-Math-7B} as our primary evaluation model. 
This choice is motivated by its moderate mathematical proficiency: it demonstrates non-trivial reasoning skills while still leaving sufficient headroom on our quiz tasks, thus offering a reliable foundation for assessing the impact of \methodname{}. 
For this model, we report results for all three \methodname{} variants (\methodname{}-Base, \methodname{}-CR, and \methodname{}-KL). 
Beyond this, we further evaluate \methodname{}-CR on \textbf{DeepSeek-R1-Distill-Qwen-1.5B}, \textbf{Qwen2.5-Math-1.5B}, and \textbf{Llama-3-8B-Instruct}, highlighting the algorithm’s robustness and cross-model generalization across both math-specialized and instruction-tuned settings.

\subsection{Training Settings}
Our \methodname{} interventions introduce method-specific hyperparameters, which are defined on top of the \methodname{}-Base configuration, that is, the original GRPO setup with generated quizzes. For the \textbf{Concept-Guided Trajectory Replacement} method, concept-guided trajectories receive a reward bonus of $r_{\text{bonus}} = 0.4$. For the \textbf{Concept-Guided KL-Regularization}  method, an \textbf{additional and dynamic} KL coefficient $\lambda_{\text{KL}}$ is applied: we use $\lambda_{\text{KL}} = 0.03$ if the reference concept-guided trajectory is correct, and $\lambda_{\text{KL}} = 0.005$ if it is incorrect. A comprehensive list of all other hyperparameters, derived from our training script, is provided in Appendix~\ref{sec:more-training-details}..\\

\subsection{Evaluation Settings}
\paragraph{In-domain Test.}
To measure in-domain performance, we use the 140 multiple-choice exercises curated from the textbook. These exercises serve as a high-quality and reliable measure of concept application due to their expert authorship and direct alignment with the textbook's definitions. We denote this test set as \textbf{Textbook}.

\paragraph{Out-of-domain Benchmarks.}
To assess whether the conceptual understanding fostered by \methodname{} generalizes beyond our curated training data, we evaluate our trained models on a diverse suite of out-of-domain benchmarks. These benchmarks were specifically chosen to probe for different facets of mathematical reasoning, from multi-step arithmetic and competition math to robustness against perturbations. This evaluation is critical to demonstrate that \methodname{} does not simply overfit to the textbook's style, but rather instills a more fundamental and transferable reasoning capability.

We evaluate models trained with three instantiations of our \methodname{} framework—\methodname{}-Base, \methodname{}-CR, and \methodname{}-KL—on the following out-of-distribution benchmarks: \textbf{GSM8K} \citep{cobbe2021gsm8k}, \textbf{ASDiv} \citep{miao2021diversecorpusevaluatingdeveloping}, \textbf{MAWPS} \citep{koncel-kedziorski-etal-2016-mawps}, \textbf{TabMWP} \citep{lu2023dynamic}, \textbf{MATH} \citep{hendrycks2021measuring}, \textbf{MMLU-STEM} \citep{hendrycks2021measuringmassivemultitasklanguage}, \textbf{Gaokao 2023 (EN)} \citep{zhong2023agieval}, 
\textbf{Gaokao-Math-QA} \citep{zhong2023agieval},
\textbf{CMATH} \citep{wei2023cmathlanguagemodelpass},
\textbf{Minerva Math} \citep{lewkowycz2022solvingquantitativereasoningproblems},
\textbf{SVAMP} \citep{patel2021nlpmodelsreallyable},
\textbf{CounterMath} \citep{li2025one}, \textbf{TheoremQA} \citep{chen-etal-2023-theoremqa}, and \textbf{OlympiadBench} \citep{he-etal-2024-olympiadbench}. A detailed description of the datasets is provided in Appendix~\ref{appendix:Evaulation Dataset}.\

\paragraph{Metrics.}
Across all benchmark evaluations in this paper, we employ a self-consistency protocol to ensure robust and stable results. For each problem, we sample 21 distinct reasoning paths by setting a high sampling temperature ($T=0.7$). We denote this as \textbf{SC@21}. More evaluation details are shown in the Appendix~\ref{sec:more-evaluation-details}


\begin{table}[htbp]
\centering
\caption{Main table of accuracy (\%) under $\mathrm{SC}@21$ ($T{=}0.7$). 
Columns use \emph{two-letter} abbreviations: \textbf{TB}=Textbook, \textbf{GS}=GSM8K, \textbf{AD}=ASDiv, 
\textbf{MW}=MAWPS, \textbf{TM}=TabMWP, \textbf{MH}=MATH, \textbf{MS}=MMLU-STEM, 
\textbf{GK}=Gaokao 2023 (EN), \textbf{CM}=CounterMath (reported as F1), \textbf{TQ}=TheoremQA, \textbf{OL}=OlympiadBench.}
\label{tab:ood_results_sc}
\setlength{\tabcolsep}{4pt}
\renewcommand{\arraystretch}{1.15}
\resizebox{\linewidth}{!}{%
\begin{tabular}{ll*{11}{c}}
\toprule
Model & Method & TB & GS & AD & MW & TM & MH & MS & GK & CM & TQ & OL \\
\midrule
\multirow{5}{*}{Qwen2-Math-7B}
& Vanilla            & 46.4 & 89.8 & 95.1 & 96.8 & 90.2 & 69.1 & 72.9 & 55.3 & 13.2  & 34.6 & 28.7 \\
& SFT                & 45.0 & 86.6 & 94.1 & 96.6 & 85.6 & 59.4 & 72.4 & 46.5 & \textbf{16.7}  & \textbf{44.2} & 19.7 \\
& \methodname{}-Base & 50.7 & 90.8 & 95.4 & 97.2 & 92.6 & 71.1 & 72.9 & \textbf{59.5} & 13.5 & 40.4 & 33.9 \\
& \methodname{}-CR   & 52.1 & \textbf{91.1} & \textbf{95.7} & 97.3 & \textbf{93.6} & \textbf{71.4} & 72.6 & 58.4 & 15.5 & 42.3 & \textbf{34.5} \\
& \methodname{}-KL   & \textbf{55.7} & 90.7 & 95.5 & \textbf{97.5} & 90.6 & 70.5 & \textbf{73.1} & \textbf{59.5} & 15.8 & \textbf{44.2} & 32.9 \\
\bottomrule
\end{tabular}%
}
\end{table}

As shown in Table~\ref{tab:ood_results_sc}, models trained with the \methodname{} framework exhibit consistent and significant performance improvements across the majority of out-of-domain benchmarks when compared to the vanilla baseline. This demonstrates that \methodname{} successfully enhances the models' underlying reasoning abilities in a way that generalizes to unseen problem distributions and formats.


\begin{table}[htbp]
\centering
\caption{DeepSeek-R1-Distill-Qwen-1.5B, Qwen2.5-Math-1.5B, and Llama-3-8B-Instruct: Out-of-domain benchmark accuracy (\%) under $\mathrm{SC}@21$ ($T{=}0.7$). 
Columns use \emph{two-letter} abbreviations: 
\textbf{CA}=CMath,
\textbf{GQ} = GaokaoMathQA,
\textbf{GK}=Gaokao 2023 (EN),
\textbf{MH}=MATH,
\textbf{MW}=MAWPS,
\textbf{MM}=Minerva Math,
\textbf{MS}=MMLU-STEM,
\textbf{SV}=SVAMP,
\textbf{TM}=TabMWP}
\label{tab:llama_ood_results_sc}
\setlength{\tabcolsep}{4pt}
\renewcommand{\arraystretch}{1.15}
\resizebox{\linewidth}{!}{%
\begin{tabular}{ll*{12}{c}}
\toprule
Model & Method & CA & GQ & GK & MH & MW & MM & MS & SV & TM \\
\midrule
\multirow{2}{*}{DeepSeek-R1-DQ-1.5B}
& Vanilla            & 90.8 & 75.2 & 58.2 & 68.6 & 96.9 & 23.9 & 58.6 & 92.8 & \textbf{89}  \\
& \methodname{}-CR  & \textbf{91.5} & \textbf{75.5} & \textbf{59.2} & \textbf{69} & \textbf{97.1} & \textbf{24.3} & \textbf{59.9} & \textbf{94} & 87.6\\
\bottomrule

\multirow{2}{*}{Qwen2.5-Math-1.5B}
& Vanilla            & \textbf{91} & \textbf{60.7} & 59.5 & 75.9 & 97.1 & 26.1 & \textbf{61.2} & 93 & 84\\
& \methodname{}-CR  & \textbf{91} & 57.8 & \textbf{60} & \textbf{77.2} & \textbf{97.6} & \textbf{29.4} & 59.4 & \textbf{93.3} & \textbf{85.9}\\
\bottomrule

\multirow{2}{*}{Llama-3-8B-Inst}
& Vanilla          & 78.8 & 25.9 & 35.8 & \textbf{41.6} & 93.8 & \textbf{16.9} & 63.2 & 90 & 77.1\\
& \methodname{}-CR  & \textbf{79.7} & \textbf{26.2} & \textbf{36.6} & 39.9 & \textbf{95.4} & 15.8 & \textbf{64.6} & \textbf{91.6} & \textbf{80.4}\\
\bottomrule
\end{tabular}%
}
\end{table}

\section{Analysis}

\subsection{Does \methodname{} Enhance Concept Selection and Application?}

\begin{table}[t]
  \centering
  \small
  \begin{tabular}{lcc}
    \toprule
    Category & \# Problems & Probability(\%) \\
    \midrule
    Concept-selection & 10 & 52.6 \\
    Mixed & 9 & 47.4 \\
    Heuristic-selection & 0 & 0.0 \\
    \midrule
    Total & 19 & 100.0 \\
    \bottomrule
  \end{tabular}
  \caption{Results on the diagnostic subset $W$ (\(|W|=19\)). A problem is Concept-Selection iff both \methodname{}-CR and \methodname{}-KL explicitly invoke the target concept and show no heuristic cues; Heuristic-selection iff both rely on heuristics with no substantive concept use; otherwise Mixed.}
  \label{tab:diag_w_results}
\end{table}

We first verify what training with \methodname{} actually changed: are the observed accuracy gains attributable to improved \emph{concept selection and application}, rather than superficial heuristics, and achieved without any test-time concept prompting? To probe this, we evaluate four model variants (Vanilla, \methodname{} Base, \methodname{} CR, and \methodname{} KL) 
on 140 textbook exercises, and define a diagnostic subset $W$ comprising problems that are solved 
by both \methodname{} CR and \methodname{} KL but missed by either Vanilla or \methodname{} Base,
\[
W=\{\,i\mid (\text{Vanilla fails on }i\ \text{or \methodname{}-BASE fails on }i)\ \land\ (\methodname{}\text{-CR and }\methodname{}\text{-KL succeed on }i)\,\},
\]
yielding $|W|=19$ (12 Vanilla-only failures, 4 \methodname{}-BASE-only failures, 3 shared failures). For each $i\in W$, we read the generations from \methodname{}-CR and \methodname{}-KL and score along two dimensions: (i) \textbf{concept hits}—the output explicitly mentions the task’s target concept and uses it correctly in the reasoning; and (ii) \textbf{heuristic cues}—guessing, option elimination without justification, surface pattern matching, or plug-in substitution without conceptual warrant.

For labeling, we keep the rules simple. A problem is called Concept-Selection if both CORE outputs (\methodname{}-CR and \methodname{}-KL) contain a concept hit and neither shows heuristic cues; It is classified as Heuristic-Selection if both outputs rely on heuristics and contain no concept hit; otherwise, it is categorized as Concept+Heuristic (mixed). We require “two hits” (one per CORE variant) so that if either variant omits the concept, the instance is not counted as Concept-Selection. As Table~\ref{tab:diag_w_results} shows, 10/19 (52.6\%) cases are Concept-Selection , 9/19 (47.4\%) are Mixed, and 0/19 are Heuristic-selection. Due to our strict constraints, it rules out superficial shortcutting as the primary driver of the gains. On one representative question (see Table \ref{tab:test_time_prompt}), even after appending a targeted concept prompt at test time, \methodname{}-Base remained incorrect, whereas \methodname{}-KL solved it without any prompt. While this is a single illustrative case, it reinforces that the observed gains come from training-induced mechanism change rather than prompt engineering. 
Taken together, the evidence indicates a mechanism shift: \methodname{} improves accuracy mainly by strengthening concept selection and application.



\subsection{Does \methodname{} Improve Robustness to Irrelevant Concept Perturbations?}

We next ask whether training with \methodname{} yields improved \emph{robustness} to irrelevant concept cues. Using 140 high-quality, thematically related textbook exercises, we prepend $K \in \{1,2,3\}$ concepts that are not directly related to the target concept to each question and measure whether the model can still retain the correct answer under such perturbations. To ensure that distractors are not directly related, we select concepts drawn from different textbook chapters. For each question and each $K$, we sample one fixed distractor set once (single random seed) and use the same set for all models to enable paired comparisons.

\begin{figure}[ht]
    \centering
    \includegraphics[width=\linewidth]{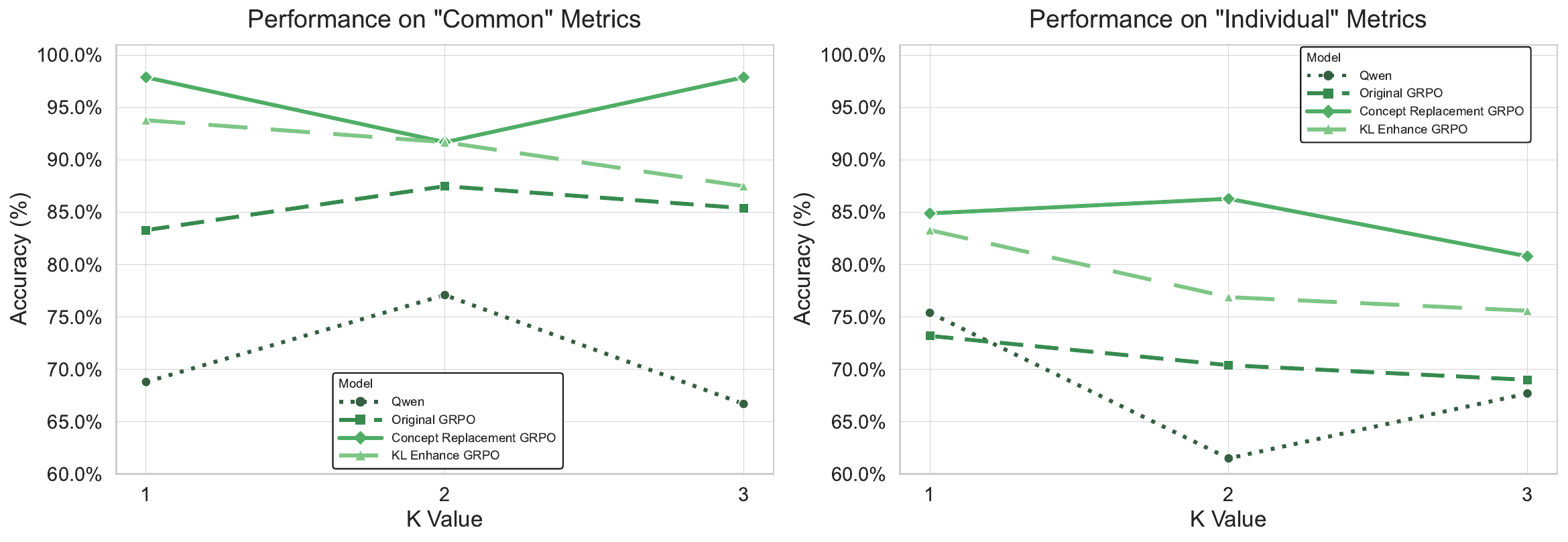}
    \caption{Performance comparison on Common vs Individual metrics.}
    \label{fig:results}
\end{figure}

To quantify retention under perturbation, we report \textbf{RUD}\textsubscript{$K$} (Retention Under Distractors): accuracy on perturbed items restricted to questions a model already solved without perturbation. Formally, letting $S_m$ be the questions solved by model $m$ in the unperturbed setting, $x_i^{(K)}$ be the question with $K$ distractors prepended and $y_i$ be the correct label,
\[
\mathrm{RUD}_K(m)\;=\;\frac{1}{|S_m|}\sum_{i\in S_m}\mathbf{1}\!\left\{\,m\!\left(x_i^{(K)}\right)=y_i\,\right\}.
\]
We evaluate four models—Vanilla, \methodname{}-Base, \methodname{}-CR, and \methodname{}-KL—under two splits: \textbf{Common} (items solved by all models; $n=48$) and \textbf{Individual} (per-model solved sets: Vanilla 65 / \methodname{}-Base 71 / \methodname{}-CR 73 / \methodname{}-KL 78). The resulting $\mathrm{RUD}_K$ curves for each split can be seen in Figure~\ref{fig:results}.

As $K$ increases, models trained with \methodname{} show consistently smaller accuracy drops than Vanilla and \methodname{}-Base on both splits, with the \methodname{}-\textsc{CR} variant particularly robust. This pattern indicates that \methodname{} not only improves the accuracy of the headline, but also improves the robustness, making predictions more stable against perturbations of irrelevant concepts.

\subsection{Does \methodname{} Apply to Base and Instruction-Tuned Models?}
We ask whether these gains persist across both base and instruction-tuned models. We therefore apply \methodname{}-CR to three representative models: \textbf{DeepSeek-R1-Distill-Qwen-1.5B}, \textbf{Qwen2.5-Math-1.5B}, and \textbf{Llama-3-8B-Instruct}. We evaluate them under the same $\mathrm{SC}@21$ ($T{=}0.7$) protocol across diverse out-of-domain suites (Table~\ref{tab:llama_ood_results_sc}; abbreviations follow the table).

\methodname{}-CR yields consistent average improvements across all three models: 
DeepSeek-R1-DQ-1.5B improves from $72.7 \rightarrow 73.1$ $(+0.4)$, 
Qwen2.5-Math-1.5B from $72.1 \rightarrow 72.4$ $(+0.3)$, 
and Llama-3-8B-Inst from $58.1 \rightarrow 58.9$ $(+0.8)$.
These results indicate that \methodname{} is \emph{model-agnostic}, providing plug-in gains without any retraining.

For DeepSeek-R1-DQ-1.5B, we observe broad gains on $\text{GK}~(+1.0)$, $\text{MS}~(+1.3)$, and $\text{SV}~(+1.2)$, with a small drop on $\text{TM}~(-1.4)$.
For Qwen2.5-Math-1.5B, the strongest improvements appear on $\text{MM}~(+3.3)$, $\text{TM}~(+1.9)$, and $\text{MH}~(+1.3)$; modest regressions occur on $\text{GQ}~(-2.9)$ and $\text{MS}~(-1.8)$.
For Llama-3-8B-Inst, \methodname{}-CR notably improves $\text{TM}~(+3.3)$, $\text{MW}~(+1.6)$, $\text{SV}~(+1.6)$, and $\text{MS}~(+1.4)$, with small drops on $\text{MH}~(-1.7)$ and $\text{MM}~(-1.1)$.

Taken together, \methodname{} complements both base and instruction-tuned models, yielding higher accuracy and stable performance across varied out-of-domain benchmarks.

\subsection{Do the Gains Come from \methodname{} Rather than GRPO?}
We aim to disentangle whether the gains arise from the textbook-style CORE curriculum introduced by \methodname{} or from artifacts of GRPO. Accordingly, we perform controlled ablations on \textbf{Qwen2-Math-7B}, holding compute, data, and hyperparameters fixed, and evaluate all variants under the same $\mathrm{SC}@21$ ($T{=}0.7$) protocol (Table~\ref{tab:RQ4}).

\textbf{Random-Reward GRPO.} For each generated response, we assign a binary reward at random (correct/incorrect) and train with GRPO on these randomized labels. This tests whether stochastic reward noise alone can yield apparent gains.

\textbf{More-Candidates Top-$k$ Control.} Standard GRPO samples four responses per input. Here we additionally generate two extra candidates (six in total) and then select the four highest-reward responses for training. This controls for the effect of a larger rollout count and stronger in-GRPO selection pressure, independent of \methodname{}.

\textbf{Results.} Table~\ref{tab:RQ4} shows that \emph{Random-Reward GRPO} does not yield meaningful gains under the same setup. The \emph{Top 4 of six} variant produces only small and unstable changes, often mild regressions. This indicates that larger rollouts or stricter within-GRPO selection by itself does not account for the improvements. By contrast, both \emph{CORE-Base} and \emph{CORE-CR} outperform the vanilla baseline across benchmarks, with \emph{CORE-CR} achieving the best overall performance.

 The results point to \methodname{} as the main source of improvement. GRPO on its own, whether with random rewards or with a larger top-$k$ selection, does not reproduce the effect; once \methodname{} is removed, the gains disappear.
\begin{table}[htbp]
\centering
\caption{Ablation on \textbf{Qwen2-Math-7B} under $\mathrm{SC}@21$ ($T{=}0.7$).
We compare: \emph{Vanilla}; \emph{\methodname{}-Base}; \emph{Random-Reward GRPO}; \emph{Top 4 of 6 GRPO}; and \emph{\methodname{}-CR}.
Metrics are accuracy (\%). Abbreviations: \textbf{AD}=\textsc{ASDiv}, \textbf{GK}=\textsc{Gaokao 2023 (EN)}, \textbf{GS}=\textsc{GSM8K}, \textbf{MW}=\textsc{MAWPS}, \textbf{OL}=\textsc{OlympiadBench}, \textbf{SV}=\textsc{SVAMP}, \textbf{TM}=\textsc{TabMWP}.}
\label{tab:RQ4}
\setlength{\tabcolsep}{4pt}
\renewcommand{\arraystretch}{1.15}
\resizebox{\linewidth}{!}{%
\begin{tabular}{ll*{9}{c}}
\toprule
Model & Method & AD & GK & GS & MW & OL & SV & TM \\
\midrule
\multirow{5}{*}{Qwen2\text{-}Math\text{-}7B}
& Vanilla                     & 95.1 & 55.3 & 89.8 & 96.8 & 28.7 & 92.5 & 90.2 \\
& \methodname{}-Base                  & 95.4 & \textbf{59.5} & 90.8 & 97.2 & 33.9 & 92.1 & 92.6 \\
& Random-Reward GRPO        & 95.2 & 55.8 & 89.9 & 97.2 & 29.9 & 91.2 & 89.6 \\
& Top 4 of 6 GRPO& 95 & 54.5 & 89.3 & \textbf{97.3} & 32.1 & 91.5 & 88.3 \\
& \methodname{}-CR     & \textbf{95.7} & 58.4 & \textbf{91.1} & \textbf{97.3} & \textbf{34.5} & \textbf{92.7} & \textbf{93.6} \\
\bottomrule
\end{tabular}%
}
\end{table}

\subsection{Is CORE Driven by Knowledge Distillation or Intrinsic Conceptual Reinforcement?}
\label{sec:intrinsic_vs_distill}

A critical concern in RL training with synthetic data is whether the observed gains stem from simple \textbf{knowledge distillation} from a superior teacher model. To disentangle CORE's mechanism from such effects, we conducted a ``Self-Supervised'' experiment by restricting the entire pipeline to the \textbf{Qwen2-Math-7B} family, thereby eliminating any external ``expert'' guidance from ultra-large LLMs.

Specifically, we utilized \textbf{Qwen2-Math-7B-Instruct} to generate 885 quizzes based on the 236 textbook concepts. We opted for the \texttt{Instruct} variant for data synthesis because, while the learner model (\textbf{Qwen2-Math-7B}) is capable of solving problems, it struggles to generate a large volume of diverse questions. To ensure quality without external supervision, we employed a self-verification protocol where only the 670 quizzes that the model could solve correctly via self-consistency (SC@21) were retained.

As shown in Table~\ref{tab:self_supervised}, \methodname{}-CR delivers consistent and substantial improvements across benchmarks even with self-generated and potentially noisy concept probes, including notable gains on GSM8K (+1.6), TabMWP (+2.3), and OlympiadBench (+4.2). This evidence provides a definitive answer: CORE's efficacy does not depend on a perfectly clean teacher or a superior expert model. Instead, the framework is \textbf{naturally robust}, capable of extracting useful learning signals through its intrinsic intervention logic, effectively bridging the reasoning gap through a closed-loop, self-improving process.

\begin{table}[t]
  \centering
  \small
  \caption{Results of the Self-Supervised experiment using \textbf{Qwen2-Math-7B-Instruct} as the generator and \textbf{Qwen2-Math-7B} as the learner. Abbreviations: \textbf{AD}=\textsc{ASDiv}, \textbf{GK}=\textsc{Gaokao 2023 (EN)}, \textbf{GS}=\textsc{GSM8K}, \textbf{MH}=\textsc{MATH}, \textbf{MS}=\textsc{MMLU-STEM}, \textbf{MW}=\textsc{MAWPS}, \textbf{OL}=\textsc{OlympiadBench}, \textbf{TM}=\textsc{TabMWP}.}
  \label{tab:self_supervised}
  \begin{tabular}{lcc}
    \toprule
    Benchmark & Vanilla & \methodname{}-CR (Self-Supervised) \\
    \midrule
    GS & 89.8 & \textbf{91.4} \\
    AD & 95.1 & \textbf{95.5} \\
    MW & 96.8 & \textbf{97.6} \\
    TM & 90.2 & \textbf{92.5} \\
    MH & 69.1 & \textbf{70.4} \\
    MS & 72.9 & \textbf{73.1} \\
    GK & 55.3 & \textbf{57.7} \\
    OL & 28.7 & \textbf{32.9} \\
    \bottomrule
  \end{tabular}
\end{table}

Additionally, a verifier-guided RL variant is evaluated, termed \textbf{Process Supervision and Verifier-Guided RL} and denoted as \textbf{\methodname{}-Base + Verifier}, to contrast verifier-based process rewards with \methodname{}.
This variant conditions learning signals on intermediate reasoning trajectories rather than solely on final outcomes.

Concretely, a \textbf{Concept-Verifier} is integrated into the \methodname{}-Base training loop, using the same \textbf{Qwen2-Math-7B} model as the verifier.
For each generated reasoning trajectory, the verifier checks whether the target concept is explicitly and correctly invoked in the reasoning process.
If the concept is correctly applied, an intrinsic process reward of $+0.4$ is assigned, independent of the final answer correctness.
This verifier therefore serves as a sparse process-level reward model targeting conceptual alignment.
All other hyperparameters are kept identical to those used in \methodname{}.

Table~\ref{tab:process_verifier} summarizes the performance of this verifier-guided RL baseline.
While \methodname{}-Base + Verifier yields modest gains over the vanilla model on several benchmarks, it consistently underperforms \textbf{\methodname{}-CR}, suggesting that explicit concept-guided intervention during training is more effective than verifier-based process supervision alone.

\begin{table}[t]
\centering
\caption{Comparison with a verifier-guided RL baseline under $\mathrm{SC}@21$ ($T{=}0.7$).
Columns use \emph{two-letter} abbreviations consistent:
\textbf{GS}=GSM8K, \textbf{AD}=ASDiv, \textbf{MW}=MAWPS, \textbf{TM}=TabMWP,
\textbf{MH}=MATH, \textbf{MS}=MMLU-STEM, \textbf{OL}=OlympiadBench.}
\label{tab:process_verifier}
\setlength{\tabcolsep}{4pt}
\renewcommand{\arraystretch}{1.15}
\resizebox{\linewidth}{!}{%
\begin{tabular}{ll*{7}{c}}
\toprule
Model & Method & GS & AD & MW & TM & MH & MS & OL \\
\midrule
\multirow{3}{*}{Qwen2-Math-7B}
& Vanilla                  & 89.8 & 95.1 & 96.8 & 90.2 & 69.1 & \textbf{72.9} & 28.7 \\
& \methodname{}-Base + Verifier
                           & 90.8 & 94.5 & 96.0 & 92.5 & 69.7 & \textbf{72.9} & 34.2 \\
& \methodname{}-CR         & \textbf{91.1} & \textbf{95.7} & \textbf{97.3} & \textbf{93.6} & \textbf{71.4} & 72.6 & \textbf{34.5} \\
\bottomrule
\end{tabular}%
}
\end{table}

\section{Conclusion}

In this work, we introduced \textbf{\methodname{} (Concept-Oriented REinforcement)}, a reinforcement learning framework designed to bridge the definition--application gap in mathematical reasoning. By curating a high-quality concept--exercise corpus, diagnosing the limits of current LLMs, and injecting explicit concept signals into training via concept-aligned quizzes, concept-guided trajectory replacement, and KL-based divergence regularization, CORE provides fine-grained supervision beyond outcome correctness.

Extensive experiments on both base and instruction-tuned models demonstrate that CORE consistently improves performance on in-domain and out-of-domain benchmarks, yielding gains in concept selection, application, and robustness under perturbations. Importantly, these improvements arise without architectural modifications and are compatible with standard policy-gradient methods, underscoring the generality of the framework.

Our findings highlight that explicitly grounding reinforcement learning in mathematical concepts can substantially enhance the reasoning capabilities of LLMs, moving them beyond surface heuristics toward genuine conceptual competence. We hope this work motivates further exploration of concept-centered training signals, not only in mathematics but also across domains where principled reasoning is essential.

\newpage
\subsubsection*{Ethics Statement}
Our dataset is curated from high-quality educational resources originally published in Chinese. We contacted the primary author, who indicated they could not grant permission at this time due to unclear regulations around LLM training and evaluation in Copyright Law of China. After consulting legal guidance, we understand that limited use of such materials for non-commercial academic research may be permissible. Accordingly, our use is strictly for research and education; we do not redistribute substantial verbatim text. The research artifacts (codes, prompts, scripts, structured concept–exercise mappings, and model-generated quizzes/snippets) are derived and only small illustrative samples, that does not contain substantial portions of the original expression, would be presented. We cite sources and will promptly honor takedown or correction requests. Any released artifacts are for research use only and may not be used commercially; parties seeking commercial use should contact the rights holders. This statement is not legal advice, and we will adjust our practices as regulations evolve.

\subsubsection*{Reproducibility Statement}
Our proposed framework is specified in \S\ref{tag:method}. Our data curation, motivation verification, and training recipes are illustraed under it with subsections named as \textit{Dataset Curation}, \textit{Gap Diagnostics}, and \textit{Concept Reinforcement Recipe}. The training and evaluation settings appear in \S\ref{sec:experiments} and Appendix \ref{appendix:Experiment}.


\bibliography{iclr2026_conference}

\begin{thebibliography}{46}
\providecommand{\natexlab}[1]{#1}
\providecommand{\url}[1]{\texttt{#1}}
\expandafter\ifx\csname urlstyle\endcsname\relax
  \providecommand{\doi}[1]{doi: #1}\else
  \providecommand{\doi}{doi: \begingroup \urlstyle{rm}\Url}\fi

\bibitem[{Anthropic}(2025)]{anthropic2025claude37}
{Anthropic}.
\newblock Claude 3.7 sonnet and claude code.
\newblock Anthropic News, 2025.
\newblock URL \url{https://www.anthropic.com/news/claude-3-7-sonnet}.
\newblock Accessed: 2025-09-22.

\bibitem[Azerbayev et~al.(2024)Azerbayev, Schoelkopf, Paster, Santos, McAleer,
  Jiang, Deng, Biderman, and Welleck]{azerbayev2024llemmaopenlanguagemodel}
Zhangir Azerbayev, Hailey Schoelkopf, Keiran Paster, Marco~Dos Santos, Stephen
  McAleer, Albert~Q. Jiang, Jia Deng, Stella Biderman, and Sean Welleck.
\newblock Llemma: An open language model for mathematics, 2024.
\newblock URL \url{https://arxiv.org/abs/2310.10631}.

\bibitem[Chen et~al.(2025)Chen, Gu, Huang, Huang, Jiang, Jie, Jin, Jin, Li, Ma,
  Ren, Shen, Shi, Sun, Sun, Wang, Wang, Wang, Wei, Wei, Wu, Wu, Xia, Xin, Yang,
  Ying, Yuan, Yuan, Zhan, Zhang, Zhang, Zhang, Zhao, Zhao, Zhou, and
  Zhu]{chen2025seedproverdeepbroadreasoning}
Luoxin Chen, Jinming Gu, Liankai Huang, Wenhao Huang, Zhicheng Jiang, Allan
  Jie, Xiaoran Jin, Xing Jin, Chenggang Li, Kaijing Ma, Cheng Ren, Jiawei Shen,
  Wenlei Shi, Tong Sun, He~Sun, Jiahui Wang, Siran Wang, Zhihong Wang, Chenrui
  Wei, Shufa Wei, Yonghui Wu, Yuchen Wu, Yihang Xia, Huajian Xin, Fan Yang,
  Huaiyuan Ying, Hongyi Yuan, Zheng Yuan, Tianyang Zhan, Chi Zhang, Yue Zhang,
  Ge~Zhang, Tianyun Zhao, Jianqiu Zhao, Yichi Zhou, and Thomas~Hanwen Zhu.
\newblock Seed-prover: Deep and broad reasoning for automated theorem proving,
  2025.
\newblock URL \url{https://arxiv.org/abs/2507.23726}.

\bibitem[Chen et~al.(2023)Chen, Yin, Ku, Lu, Wan, Ma, Xu, Wang, and
  Xia]{chen-etal-2023-theoremqa}
Wenhu Chen, Ming Yin, Max Ku, Pan Lu, Yixin Wan, Xueguang Ma, Jianyu Xu, Xinyi
  Wang, and Tony Xia.
\newblock {T}heorem{QA}: A theorem-driven question answering dataset.
\newblock In Houda Bouamor, Juan Pino, and Kalika Bali (eds.),
  \emph{Proceedings of the 2023 Conference on Empirical Methods in Natural
  Language Processing}, pp.\  7889--7901, Singapore, December 2023. Association
  for Computational Linguistics.
\newblock \doi{10.18653/v1/2023.emnlp-main.489}.
\newblock URL \url{https://aclanthology.org/2023.emnlp-main.489/}.

\bibitem[Cobbe et~al.(2021)Cobbe, Kosaraju, Bavarian, Chen, Jun, Kaiser,
  Plappert, Tworek, Hilton, Nakano, Hesse, and Schulman]{cobbe2021gsm8k}
Karl Cobbe, Vineet Kosaraju, Mohammad Bavarian, Mark Chen, Heewoo Jun, Lukasz
  Kaiser, Matthias Plappert, Jerry Tworek, Jacob Hilton, Reiichiro Nakano,
  Christopher Hesse, and John Schulman.
\newblock Training verifiers to solve math word problems.
\newblock \emph{arXiv preprint arXiv:2110.14168}, 2021.

\bibitem[DeepSeek-AI(2025{\natexlab{a}})]{deepseekai2025deepseekr1incentivizingreasoningcapability}
DeepSeek-AI.
\newblock Deepseek-r1: Incentivizing reasoning capability in llms via
  reinforcement learning, 2025{\natexlab{a}}.
\newblock URL \url{https://arxiv.org/abs/2501.12948}.

\bibitem[DeepSeek-AI(2025{\natexlab{b}})]{deepseekai2025deepseekv3technicalreport}
DeepSeek-AI.
\newblock Deepseek-v3 technical report, 2025{\natexlab{b}}.
\newblock URL \url{https://arxiv.org/abs/2412.19437}.

\bibitem[Feng et~al.(2025)Feng, Zhou, Lin, and
  Roth]{feng2025birdtrustworthybayesianinference}
Yu~Feng, Ben Zhou, Weidong Lin, and Dan Roth.
\newblock Bird: A trustworthy bayesian inference framework for large language
  models, 2025.
\newblock URL \url{https://arxiv.org/abs/2404.12494}.

\bibitem[Gemini(2025)]{comanici2025gemini25pushingfrontier}
Team Gemini.
\newblock Gemini 2.5: Pushing the frontier with advanced reasoning,
  multimodality, long context, and next generation agentic capabilities, 2025.
\newblock URL \url{https://arxiv.org/abs/2507.06261}.

\bibitem[Guo et~al.(2025{\natexlab{a}})Guo, Liu, Fan, He, Li, Wang, and
  Fung]{guo2025mathematicalprooflitmustest}
Dadi Guo, Jiayu Liu, Zhiyuan Fan, Zhitao He, Haoran Li, Yumeng Wang, and Yi~R.
  Fung.
\newblock Mathematical proof as a litmus test: Revealing failure modes of
  advanced large reasoning models, 2025{\natexlab{a}}.
\newblock URL \url{https://arxiv.org/abs/2506.17114}.

\bibitem[Guo et~al.(2025{\natexlab{b}})Guo, Yang, Zhang, Song, Zhang, Xu, Zhu,
  Ma, Wang, Bi, et~al.]{deepseek2025r1}
Daya Guo, Dejian Yang, Haowei Zhang, Junxiao Song, Ruoyu Zhang, Runxin Xu,
  Qihao Zhu, Shirong Ma, Peiyi Wang, Xiao Bi, et~al.
\newblock Deepseek-r1: Incentivizing reasoning capability in llms via
  reinforcement learning.
\newblock \emph{arXiv preprint arXiv:2501.12948}, 2025{\natexlab{b}}.
\newblock URL \url{https://arxiv.org/abs/2501.12948}.

\bibitem[Guo et~al.(2025{\natexlab{c}})Guo, Yang, Zhang, Xu, Du, Zheng, and
  Huang]{guo2025rightenoughpitfallsoutcome}
Jiaxing Guo, Wenjie Yang, Shengzhong Zhang, Tongshan Xu, Lun Du, Da~Zheng, and
  Zengfeng Huang.
\newblock Right is not enough: The pitfalls of outcome supervision in training
  llms for math reasoning, 2025{\natexlab{c}}.
\newblock URL \url{https://arxiv.org/abs/2506.06877}.

\bibitem[He et~al.(2024)He, Luo, Bai, Hu, Thai, Shen, Hu, Han, Huang, Zhang,
  Liu, Qi, Liu, and Sun]{he-etal-2024-olympiadbench}
Chaoqun He, Renjie Luo, Yuzhuo Bai, Shengding Hu, Zhen Thai, Junhao Shen, Jinyi
  Hu, Xu~Han, Yujie Huang, Yuxiang Zhang, Jie Liu, Lei Qi, Zhiyuan Liu, and
  Maosong Sun.
\newblock {O}lympiad{B}ench: A challenging benchmark for promoting {AGI} with
  olympiad-level bilingual multimodal scientific problems.
\newblock In Lun-Wei Ku, Andre Martins, and Vivek Srikumar (eds.),
  \emph{Proceedings of the 62nd Annual Meeting of the Association for
  Computational Linguistics (Volume 1: Long Papers)}, pp.\  3828--3850,
  Bangkok, Thailand, August 2024. Association for Computational Linguistics.
\newblock \doi{10.18653/v1/2024.acl-long.211}.
\newblock URL \url{https://aclanthology.org/2024.acl-long.211/}.

\bibitem[Hendrycks et~al.(2021{\natexlab{a}})Hendrycks, Burns, Basart, Zou,
  Mazeika, Song, and
  Steinhardt]{hendrycks2021measuringmassivemultitasklanguage}
Dan Hendrycks, Collin Burns, Steven Basart, Andy Zou, Mantas Mazeika, Dawn
  Song, and Jacob Steinhardt.
\newblock Measuring massive multitask language understanding,
  2021{\natexlab{a}}.
\newblock URL \url{https://arxiv.org/abs/2009.03300}.

\bibitem[Hendrycks et~al.(2021{\natexlab{b}})Hendrycks, Burns, Kadavath, Arora,
  Basart, Tang, Song, and Steinhardt]{hendrycks2021measuring}
Dan Hendrycks, Collin Burns, Saurav Kadavath, Akul Arora, Steven Basart, Eric
  Tang, Dawn Song, and Jacob Steinhardt.
\newblock Measuring mathematical problem solving with the {MATH} dataset.
\newblock In \emph{Thirty-fifth Conference on Neural Information Processing
  Systems Datasets and Benchmarks Track (Round 2)}, 2021{\natexlab{b}}.
\newblock URL \url{https://openreview.net/forum?id=7Bywt2mQsCe}.

\bibitem[Huang et~al.(2025)Huang, Guo, Li, Ji, Ge, Li, Guo, Cai, Yuan, Wang,
  Wu, Yin, Tang, Huang, Jin, Chen, Zhang, and Wang]{huang2025mathperturb}
Kaixuan Huang, Jiacheng Guo, Zihao Li, Xiang Ji, Jiawei Ge, Wenzhe Li, Yingqing
  Guo, Tianle Cai, Hui Yuan, Runzhe Wang, Yue Wu, Ming Yin, Shange Tang,
  Yangsibo Huang, Chi Jin, Xinyun Chen, Chiyuan Zhang, and Mengdi Wang.
\newblock {MATH}-perturb: Benchmarking {LLM}s' math reasoning abilities against
  hard perturbations.
\newblock In \emph{Forty-second International Conference on Machine Learning},
  2025.
\newblock URL \url{https://openreview.net/forum?id=OZy70UggXr}.

\bibitem[Huang \& Yang(2025)Huang and Yang]{huang2025gemini25procapable}
Yichen Huang and Lin~F. Yang.
\newblock Gemini 2.5 pro capable of winning gold at imo 2025, 2025.
\newblock URL \url{https://arxiv.org/abs/2507.15855}.

\bibitem[Koncel-Kedziorski et~al.(2016)Koncel-Kedziorski, Roy, Amini, Kushman,
  and Hajishirzi]{koncel-kedziorski-etal-2016-mawps}
Rik Koncel-Kedziorski, Subhro Roy, Aida Amini, Nate Kushman, and Hannaneh
  Hajishirzi.
\newblock {MAWPS}: A math word problem repository.
\newblock In Kevin Knight, Ani Nenkova, and Owen Rambow (eds.),
  \emph{Proceedings of the 2016 Conference of the North {A}merican Chapter of
  the Association for Computational Linguistics: Human Language Technologies},
  pp.\  1152--1157, San Diego, California, June 2016. Association for
  Computational Linguistics.
\newblock \doi{10.18653/v1/N16-1136}.
\newblock URL \url{https://aclanthology.org/N16-1136/}.

\bibitem[Lewkowycz et~al.(2022)Lewkowycz, Andreassen, Dohan, Dyer, Michalewski,
  Ramasesh, Slone, Anil, Schlag, Gutman-Solo, Wu, Neyshabur, Gur-Ari, and
  Misra]{lewkowycz2022solvingquantitativereasoningproblems}
Aitor Lewkowycz, Anders Andreassen, David Dohan, Ethan Dyer, Henryk
  Michalewski, Vinay Ramasesh, Ambrose Slone, Cem Anil, Imanol Schlag, Theo
  Gutman-Solo, Yuhuai Wu, Behnam Neyshabur, Guy Gur-Ari, and Vedant Misra.
\newblock Solving quantitative reasoning problems with language models, 2022.
\newblock URL \url{https://arxiv.org/abs/2206.14858}.

\bibitem[Li et~al.(2025)Li, Kuang, Huang, Xu, Liang, Yu, Lu, Li, Tan, Qu, Shen,
  Zheng, and Yu]{li2025one}
Yinghui Li, Jiayi Kuang, Haojing Huang, Zhikun Xu, Xinnian Liang, Yi~Yu,
  Wenlian Lu, Yangning Li, Xiaoyu Tan, Chao Qu, Ying Shen, Hai-Tao Zheng, and
  Philip~S. Yu.
\newblock One example shown, many concepts known! counterexample-driven
  conceptual reasoning in mathematical {LLM}s.
\newblock In \emph{Forty-second International Conference on Machine Learning},
  2025.
\newblock URL \url{https://openreview.net/forum?id=A31Ep22iQ7}.

\bibitem[Lu et~al.(2023)Lu, Qiu, Chang, Wu, Zhu, Rajpurohit, Clark, and
  Kalyan]{lu2023dynamic}
Pan Lu, Liang Qiu, Kai-Wei Chang, Ying~Nian Wu, Song-Chun Zhu, Tanmay
  Rajpurohit, Peter Clark, and Ashwin Kalyan.
\newblock Dynamic prompt learning via policy gradient for semi-structured
  mathematical reasoning.
\newblock In \emph{The Eleventh International Conference on Learning
  Representations}, 2023.
\newblock URL \url{https://openreview.net/forum?id=DHyHRBwJUTN}.

\bibitem[Luo et~al.(2025)Luo, Sun, Xu, Zhao, Lou, Tao, Geng, Lin, Chen, Tang,
  and Zhang]{luo2025wizardmathempoweringmathematicalreasoning}
Haipeng Luo, Qingfeng Sun, Can Xu, Pu~Zhao, Jianguang Lou, Chongyang Tao, Xiubo
  Geng, Qingwei Lin, Shifeng Chen, Yansong Tang, and Dongmei Zhang.
\newblock Wizardmath: Empowering mathematical reasoning for large language
  models via reinforced evol-instruct, 2025.
\newblock URL \url{https://arxiv.org/abs/2308.09583}.

\bibitem[Miao et~al.(2021)Miao, Liang, and
  Su]{miao2021diversecorpusevaluatingdeveloping}
Shen-Yun Miao, Chao-Chun Liang, and Keh-Yih Su.
\newblock A diverse corpus for evaluating and developing english math word
  problem solvers, 2021.
\newblock URL \url{https://arxiv.org/abs/2106.15772}.

\bibitem[Mirzadeh et~al.(2025)Mirzadeh, Alizadeh, Shahrokhi, Tuzel, Bengio, and
  Farajtabar]{mirzadeh2025gsmsymbolicunderstandinglimitationsmathematical}
Iman Mirzadeh, Keivan Alizadeh, Hooman Shahrokhi, Oncel Tuzel, Samy Bengio, and
  Mehrdad Farajtabar.
\newblock Gsm-symbolic: Understanding the limitations of mathematical reasoning
  in large language models, 2025.
\newblock URL \url{https://arxiv.org/abs/2410.05229}.

\bibitem[OpenAI(2024)]{openai2024openaio1card}
OpenAI.
\newblock Openai o1 system card, 2024.
\newblock URL \url{https://arxiv.org/abs/2412.16720}.

\bibitem[OpenAI(2025)]{openai2025o3}
OpenAI.
\newblock Openai o3 and o4-mini system card.
\newblock OpenAI, 2025.
\newblock URL
  \url{https://cdn.openai.com/pdf/2221c875-02dc-4789-800b-e7758f3722c1/o3-and-o4-mini-system-card.pdf}.
\newblock Accessed: 2025-09-22.

\bibitem[Patel et~al.(2021{\natexlab{a}})Patel, Bhattamishra, and
  Goyal]{patel-etal-2021-nlp}
Arkil Patel, Satwik Bhattamishra, and Navin Goyal.
\newblock Are {NLP} models really able to solve simple math word problems?
\newblock In Kristina Toutanova, Anna Rumshisky, Luke Zettlemoyer, Dilek
  Hakkani-Tur, Iz~Beltagy, Steven Bethard, Ryan Cotterell, Tanmoy Chakraborty,
  and Yichao Zhou (eds.), \emph{Proceedings of the 2021 Conference of the North
  American Chapter of the Association for Computational Linguistics: Human
  Language Technologies}, pp.\  2080--2094, Online, June 2021{\natexlab{a}}.
  Association for Computational Linguistics.
\newblock \doi{10.18653/v1/2021.naacl-main.168}.
\newblock URL \url{https://aclanthology.org/2021.naacl-main.168/}.

\bibitem[Patel et~al.(2021{\natexlab{b}})Patel, Bhattamishra, and
  Goyal]{patel2021nlpmodelsreallyable}
Arkil Patel, Satwik Bhattamishra, and Navin Goyal.
\newblock Are nlp models really able to solve simple math word problems?,
  2021{\natexlab{b}}.
\newblock URL \url{https://arxiv.org/abs/2103.07191}.

\bibitem[Qin et~al.(2025)Qin, Park, Kwun, Walsman, Malach, Anand, Tanaka, and
  Alvarez-Melis]{qin2025decomposingelementsproblemsolving}
Tian Qin, Core~Francisco Park, Mujin Kwun, Aaron Walsman, Eran Malach, Nikhil
  Anand, Hidenori Tanaka, and David Alvarez-Melis.
\newblock Decomposing elements of problem solving: What "math" does rl teach?,
  2025.
\newblock URL \url{https://arxiv.org/abs/2505.22756}.

\bibitem[Schulman et~al.(2017)Schulman, Wolski, Dhariwal, Radford, and
  Klimov]{schulman2017proximalpolicyoptimizationalgorithms}
John Schulman, Filip Wolski, Prafulla Dhariwal, Alec Radford, and Oleg Klimov.
\newblock Proximal policy optimization algorithms, 2017.
\newblock URL \url{https://arxiv.org/abs/1707.06347}.

\bibitem[Shao et~al.(2024)Shao, Wang, Zhu, Xu, Song, Bi, Zhang, Zhang, Li, Wu,
  and Guo]{shao2024deepseekmathpushinglimitsmathematical}
Zhihong Shao, Peiyi Wang, Qihao Zhu, Runxin Xu, Junxiao Song, Xiao Bi, Haowei
  Zhang, Mingchuan Zhang, Y.~K. Li, Y.~Wu, and Daya Guo.
\newblock Deepseekmath: Pushing the limits of mathematical reasoning in open
  language models, 2024.
\newblock URL \url{https://arxiv.org/abs/2402.03300}.

\bibitem[Sheng et~al.(2024)Sheng, Zhang, Ye, Wu, Zhang, Zhang, Peng, Lin, and
  Wu]{sheng2024hybridflow}
Guangming Sheng, Chi Zhang, Zilingfeng Ye, Xibin Wu, Wang Zhang, Ru~Zhang,
  Yanghua Peng, Haibin Lin, and Chuan Wu.
\newblock Hybridflow: A flexible and efficient rlhf framework.
\newblock \emph{arXiv preprint arXiv: 2409.19256}, 2024.
\newblock URL \url{https://arxiv.org/abs/2409.19256}.

\bibitem[Simon(2011)]{simon2011studying}
Martin~A Simon.
\newblock Studying mathematics conceptual learning: Student learning through
  their mathematical activity.
\newblock \emph{North American Chapter of the International Group for the
  Psychology of Mathematics Education}, 2011.
\newblock URL \url{https://eric.ed.gov/?id=ED585966}.

\bibitem[Team(2024)]{grattafiori2024llama3herdmodels}
LLaMA Team.
\newblock The llama 3 herd of models, 2024.
\newblock URL \url{https://arxiv.org/abs/2407.21783}.

\bibitem[Wei et~al.(2023)Wei, Luan, Liu, Dong, and
  Wang]{wei2023cmathlanguagemodelpass}
Tianwen Wei, Jian Luan, Wei Liu, Shuang Dong, and Bin Wang.
\newblock Cmath: Can your language model pass chinese elementary school math
  test?, 2023.
\newblock URL \url{https://arxiv.org/abs/2306.16636}.

\bibitem[Wu et~al.(2025)Wu, Zhang, Dong, Xi, Zhao, Jin, Fan, Zhou, Lv, Zhang,
  Fu, Liu, Zhang, and Zhang]{wu2025reasoningmemorizationunreliableresults}
Mingqi Wu, Zhihao Zhang, Qiaole Dong, Zhiheng Xi, Jun Zhao, Senjie Jin, Xiaoran
  Fan, Yuhao Zhou, Huijie Lv, Ming Zhang, Yanwei Fu, Qin Liu, Songyang Zhang,
  and Qi~Zhang.
\newblock Reasoning or memorization? unreliable results of reinforcement
  learning due to data contamination, 2025.
\newblock URL \url{https://arxiv.org/abs/2507.10532}.

\bibitem[Yang et~al.(2024{\natexlab{a}})Yang, Yang, Hui, Zheng, Yu, Zhou, Li,
  Li, Liu, Huang, Dong, Wei, Lin, Tang, Wang, Yang, Tu, Zhang, Ma, Yang, Xu,
  Zhou, Bai, He, Lin, Dang, Lu, Chen, Yang, Li, Xue, Ni, Zhang, Wang, Peng,
  Men, Gao, Lin, Wang, Bai, Tan, Zhu, Li, Liu, Ge, Deng, Zhou, Ren, Zhang, Wei,
  Ren, Liu, Fan, Yao, Zhang, Wan, Chu, Liu, Cui, Zhang, Guo, and
  Fan]{yang2024qwen2technicalreport}
An~Yang, Baosong Yang, Binyuan Hui, Bo~Zheng, Bowen Yu, Chang Zhou, Chengpeng
  Li, Chengyuan Li, Dayiheng Liu, Fei Huang, Guanting Dong, Haoran Wei, Huan
  Lin, Jialong Tang, Jialin Wang, Jian Yang, Jianhong Tu, Jianwei Zhang,
  Jianxin Ma, Jianxin Yang, Jin Xu, Jingren Zhou, Jinze Bai, Jinzheng He,
  Junyang Lin, Kai Dang, Keming Lu, Keqin Chen, Kexin Yang, Mei Li, Mingfeng
  Xue, Na~Ni, Pei Zhang, Peng Wang, Ru~Peng, Rui Men, Ruize Gao, Runji Lin,
  Shijie Wang, Shuai Bai, Sinan Tan, Tianhang Zhu, Tianhao Li, Tianyu Liu,
  Wenbin Ge, Xiaodong Deng, Xiaohuan Zhou, Xingzhang Ren, Xinyu Zhang, Xipin
  Wei, Xuancheng Ren, Xuejing Liu, Yang Fan, Yang Yao, Yichang Zhang, Yu~Wan,
  Yunfei Chu, Yuqiong Liu, Zeyu Cui, Zhenru Zhang, Zhifang Guo, and Zhihao Fan.
\newblock Qwen2 technical report, 2024{\natexlab{a}}.
\newblock URL \url{https://arxiv.org/abs/2407.10671}.

\bibitem[Yang et~al.(2024{\natexlab{b}})Yang, Zhang, Hui, Gao, Yu, Li, Liu, Tu,
  Zhou, Lin, Lu, Xue, Lin, Liu, Ren, and
  Zhang]{yang2024qwen25mathtechnicalreportmathematical}
An~Yang, Beichen Zhang, Binyuan Hui, Bofei Gao, Bowen Yu, Chengpeng Li,
  Dayiheng Liu, Jianhong Tu, Jingren Zhou, Junyang Lin, Keming Lu, Mingfeng
  Xue, Runji Lin, Tianyu Liu, Xingzhang Ren, and Zhenru Zhang.
\newblock Qwen2.5-math technical report: Toward mathematical expert model via
  self-improvement, 2024{\natexlab{b}}.
\newblock URL \url{https://arxiv.org/abs/2409.12122}.

\bibitem[Yao \& Xie(2015)Yao and Xie]{fudanalgebra}
Musheng Yao and Qihong Xie.
\newblock \emph{Advanced Algebra (Chinese 3rd Edition, Exercise)}.
\newblock Fudan University Press, 2015.
\newblock ISBN 9787309117769.
\newblock URL
  \url{https://www.google.com/books/edition/%E9%AB%98%E7%AD%89%E4%BB%A3%E6%95%B0/cyICkAEACAAJ?hl=en_US}.

\bibitem[Ying et~al.(2024)Ying, Zhang, Li, Zhou, Shao, Fei, Ma, Hong, Liu,
  Wang, Wang, Wu, Li, Zhou, Liu, Zhang, Zhang, Yan, Qiu, Wang, Chen, and
  Lin]{ying2024internlmmath}
Huaiyuan Ying, Shuo Zhang, Linyang Li, Zhejian Zhou, Yunfan Shao, Zhaoye Fei,
  Yichuan Ma, Jiawei Hong, Kuikun Liu, Ziyi Wang, Yudong Wang, Zijian Wu,
  Shuaibin Li, Fengzhe Zhou, Hongwei Liu, Songyang Zhang, Wenwei Zhang, Hang
  Yan, Xipeng Qiu, Jiayu Wang, Kai Chen, and Dahua Lin.
\newblock Internlm-math: Open math large language models toward verifiable
  reasoning, 2024.

\bibitem[Yu et~al.(2024)Yu, Zhou, Cheng, and
  Roth]{yu2024reasonagainusingextractablesymbolic}
Xiaodong Yu, Ben Zhou, Hao Cheng, and Dan Roth.
\newblock Reasonagain: Using extractable symbolic programs to evaluate
  mathematical reasoning, 2024.
\newblock URL \url{https://arxiv.org/abs/2410.19056}.

\bibitem[Yue et~al.(2024)Yue, Qu, Zhang, Fu, Huang, Sun, Su, and
  Chen]{yue2023mammothbuildingmathgeneralist}
Xiang Yue, Xingwei Qu, Ge~Zhang, Yao Fu, Wenhao Huang, Huan Sun, Yu~Su, and
  Wenhu Chen.
\newblock {MA}mmo{TH}: Building math generalist models through hybrid
  instruction tuning.
\newblock In \emph{The Twelfth International Conference on Learning
  Representations}, 2024.
\newblock URL \url{https://openreview.net/forum?id=yLClGs770I}.

\bibitem[Zhang et~al.(2025)Zhang, Huang, Li, Ni, Chen, and
  Oymak]{zhang2025breadbranchedrolloutsexpert}
Xuechen Zhang, Zijian Huang, Yingcong Li, Chenshun Ni, Jiasi Chen, and Samet
  Oymak.
\newblock Bread: Branched rollouts from expert anchors bridge sft \& rl for
  reasoning, 2025.
\newblock URL \url{https://arxiv.org/abs/2506.17211}.

\bibitem[Zhong et~al.(2023)Zhong, Cui, Guo, Liang, Lu, Wang, Saied, Chen, and
  Duan]{zhong2023agieval}
Wanjun Zhong, Ruixiang Cui, Yiduo Guo, Yaobo Liang, Shuai Lu, Yanlin Wang, Amin
  Saied, Weizhu Chen, and Nan Duan.
\newblock Agieval: A human-centric benchmark for evaluating foundation models,
  2023.

\bibitem[Zhou et~al.(2024)Zhou, Zhang, Chen, Yu, Wang, Peng, Roth, and
  Yu]{zhou2024conceptualunbiasedreasoninglanguage}
Ben Zhou, Hongming Zhang, Sihao Chen, Dian Yu, Hongwei Wang, Baolin Peng, Dan
  Roth, and Dong Yu.
\newblock Conceptual and unbiased reasoning in language models, 2024.
\newblock URL \url{https://arxiv.org/abs/2404.00205}.

\bibitem[Zhou et~al.(2025)Zhou, Jain, Zhang, Ning, Wang, Benajiba, and
  Roth]{zhou2025selfsupervisedanalogicallearningusing}
Ben Zhou, Sarthak Jain, Yi~Zhang, Qiang Ning, Shuai Wang, Yassine Benajiba, and
  Dan Roth.
\newblock Self-supervised analogical learning using language models, 2025.
\newblock URL \url{https://arxiv.org/abs/2502.00996}.

\end{thebibliography}
\bibliographystyle{iclr2026_conference}

\appendix
\section{Details for textbook data}
\label{appendix:data}
\subsection{Data Curation Details}

Our data curation followed a multi-stage pipeline to ensure high fidelity. We first employed an OCR tool\footnote{\url{https://github.com/opendatalab/MinerU}} to digitize the textbook. The concept and exercise sections then underwent a manual verification stage, with any recognition errors corrected using GPT-4o. Subsequently, the entire Chinese 
corpus was translated into English via GPT-4o, followed by another round of human verification on the key 
sections to ensure accuracy. 

\section{Experiment Setup}
\label{appendix:Experiment}
\subsection{More Training Details}
\label{sec:more-training-details}
In the main experiments, we train models for 3 epochs on four H100 GPUs with GRPO by the \textit{verl} \citep{sheng2024hybridflow} framework. The policy is optimized using Adam with an actor learning rate of $1 \times 10^{-6}$, and a standard KL-divergence penalty with a fixed coefficient of 0.001 is applied against the reference policy. During training, we set the sampling temperature to 0.7, with a batch size of 128 and a mini-batch size of 32. For each prompt, four responses are generated to conduct GRPO updates. Rewards follow a binary scheme (1 for correct, 0 for incorrect). The maximum prompt length is capped at 1024, while the maximum response length varies across models, being set to 1024 for Qwen2-Math-7B, 2048 for Qwen2.5-Math-1.5B and Llama-3-8B-Instruct, and 6000 for DeepSeek-R1-Distill-Qwen-1.5B to accommodate their extended reasoning contexts.

\subsection{Details of Evaulation Datasets}
\label{appendix:Evaulation Dataset}
All evaluation datasets are sourced from the Qwen2.5-Math evaluation repository\footnote{\url{https://github.com/QwenLM/Qwen2.5-Math}}. Their details are summarized below.

\subsection*{Grade School \& Middle School Level}
\begin{itemize}
    \item GSM8k: A dataset of approximately 8,500 high-quality, linguistically diverse elementary school math word problems. These problems require 2 to 8 steps of reasoning to solve and primarily involve basic arithmetic operations ($+$, $-$, $\times$, $\div$). The purpose of this dataset is to evaluate a model's ability to perform multi-step mathematical reasoning.
    \item ASDiv: An English math word problem dataset that is diverse in both linguistic patterns and problem types. It aims to comprehensively evaluate the true capabilities of math problem solvers, preventing models from achieving high scores merely by "memorizing" solutions to similar problems. Each problem is annotated with its problem type and grade level.
    \item MAWPS: A collection of several thousand English math word problems sourced from various online educational websites. Its goal is to provide an extensible repository of math problems for researchers to use and expand upon, covering a variety of basic arithmetic and algebraic problems.
    \item TabMWP: A large-scale dataset containing over 38,000 math word problems, distinguished by the inclusion of a table as context for each problem. To solve these, a model must be able to retrieve, integrate, and perform multi-step mathematical reasoning on information from both textual and tabular sources.
    \item CMath: A Chinese-language dataset designed to evaluate language models on elementary school mathematics. It contains carefully curated word problems covering fundamental arithmetic operations (addition, subtraction, multiplication, and division) and simple logic-based reasoning. The dataset challenges models to perform precise quantitative reasoning in Chinese, testing both mathematical competence and cross-lingual generalization.
    \item SVAMP: A benchmark consisting of simple arithmetic word problems created by perturbing examples from the well-known MAWPS dataset. It is designed to assess the robustness and true reasoning ability of language models by introducing variations that discourage rote memorization. Models must comprehend problem semantics and adapt to diverse formulations of similar mathematical tasks.

\end{itemize}

\subsection*{High School Level}
\begin{itemize}
    \item MATH: A dataset created by Dan Hendrycks et al., containing 12,500 problems from American high school math competitions (such as AMC 10, AMC 12, AIME). The problems cover multiple subjects including algebra, geometry, number theory, and counting \& probability. Each problem includes a detailed solution written by a human expert in \LaTeX{} format. Its difficulty is significantly higher than elementary school problems, making it a key benchmark for advanced mathematical reasoning.
    \item MMLU-STEM: MMLU includes 57 different subjects, and MMLU-STEM refers to the subset of subjects related to STEM (Science, Technology, Engineering, and Mathematics), such as college-level math, physics, chemistry, and computer science.
    \item Gaokao2023En: This dataset is derived from China's "Gaokao" (National College Entrance Examination) mathematics papers. It typically involves translating Chinese math problems into English to test a large model's ability to solve difficult math problems from different cultural and educational backgrounds.
    \item GaokaoMathQA: A high-school level benchmark constructed from authentic Chinese National College Entrance Examination (Gaokao) mathematics questions. The dataset includes both multiple-choice and open-ended problems covering a broad range of high-school topics, such as functions, geometry, probability, and calculus. It aims to evaluate models’ abilities to perform symbolic reasoning and multi-step quantitative problem-solving under exam-style constraints.

\end{itemize}

\subsection*{College Level \& Beyond}
\begin{itemize}
    \item CounterMath: A university-level mathematical benchmark designed to evaluate a model's conceptual reasoning by requiring it to prove or disprove statements by providing counterexamples. It focuses on advanced topics in Algebra, Topology, Real Analysis, and Functional Analysis.
    \item TheoremQA: A theorem-driven question answering dataset created to evaluate an AI model's ability to apply scientific theorems to solve challenging problems. It contains 800 questions covering over 350 theorems from Mathematics, Physics, EE\&CS, and Finance.
    \item Minerva Math: A benchmark derived from the work on the Minerva model \citep{lewkowycz2022solvingquantitativereasoningproblems}, which was designed to train large language models on scientific and mathematical content to enable step‐by‐step quantitative reasoning. The benchmark covers undergraduate-level math and science questions in natural language and LaTeX, requiring models to correctly parse, compute, and symbolically manipulate expressions without external tools.

\end{itemize}

\subsection*{Competition Level}
\begin{itemize}
    \item Olympiad Bench: A benchmark of extremely challenging, Olympiad-level scientific problems in both mathematics and physics. It is designed to push the boundaries of AGI research and often includes multimodal elements, requiring models to interpret diagrams and perform complex, creative reasoning.
\end{itemize}

\subsection{More Evaluation Details}
\label{sec:more-evaluation-details}
For evaluation, we adopt the SC@21 setting with a sampling temperature of 0.7. Specifically, 21 responses are generated, and after discarding cases where the answer cannot be extracted, the final prediction is determined by majority voting. In the event of a tie, one of the tied candidates is selected uniformly at random.


\subsection{Training prompt example}
\begin{tcolorbox}[
  title=\textit{Completion Prompt},
  top=1mm,
  bottom=1mm,
  colback=gray!10, 
  colframe=gray!50,
  coltitle=black
]
\footnotesize
\textbf{System:} You are a helpful assistant that solves multiple-choice math questions with step-by-step reasoning.\\[2mm]
\textbf{User:} Please solve the following question carefully. Explain your reasoning, and conclude with the final answer using the format: \verb|\boxed{X}|, where X is A, B, C, or D.\\[2mm]
Example:\\
Question: What is 2 + 3?\\
A. 4\\
B. 5\\
C. 6\\
D. 7\\[1mm]
Answer: 2 + 3 = 5, which is option B.\\
The final answer is \verb|\boxed{B}|.\\[2mm]
\textbf{---}\\
Question: \{specific math problem\}\\
A. \{option A\}\\
B. \{option B\}\\
C. \{option C\}\\
D. \{option D\}\\
\end{tcolorbox}

\newpage
\section{Analysis}
\subsection{PPO-based Variant: Optimizer-Agnostic Instantiation of \methodname{}}
\label{sec:appendix-c1}

The \methodname{} framework is designed as an architectural module that operates independently of the underlying reinforcement learning optimizer. While the primary experiments utilize GRPO for efficiency, \methodname{} can be instantiated with other policy gradient methods, such as PPO, provided the structural requirement of multi-sampling is met. This section provides the technical details of the PPO-based implementation and its corresponding performance.

\paragraph{Algorithmic Formulation}
To implement \methodname{} within a PPO backbone, a batch configuration with $N=4$ trajectories per prompt is utilized. To isolate the effects of the conceptual intervention and reduce reliance on additional neural components, the standard critic network is omitted. Instead, a group-wise Monte-Carlo baseline is employed for variance reduction. For each trajectory $i$, the discounted returns $G_i$ are first computed as follows:
\begin{equation}
G_i = \sum_{t'=t}^{T} \gamma^{t'-t} r_{t'}
\end{equation}
The advantage $A_i$ is then estimated through group normalization of these returns within the sampled group:
\begin{equation}
A_i = \frac{G_i - \mathbb{E}_{\text{group}}[G]}{\mathrm{std}_{\text{group}}(G)}
\end{equation}
This formulation facilitates the integration of the \methodname{} intervention logic, which encompasses both the Concept Bonus and Trajectory Replacement operations. Furthermore, such integration is achieved within the PPO clipped objective without necessitating any modifications to the core optimization step.

\paragraph{Experimental Results}
As summarized in Table~\ref{tab:ppo_variant}, the PPO-based instantiation of \methodname{} achieves consistent performance gains across most mathematical benchmarks. These results highlight the modularity of the framework: since \methodname{} reshapes the training distribution and reward signals at the data-processing level, it functions effectively as a higher-level supervision layer that is agnostic to the specific choice of the RL optimizer.

\begin{table}[H]
\centering
\small
\begin{tabular}{lccr}
\toprule
Benchmark & Vanilla & \methodname{}-CR (PPO) & Improvement \\
\midrule
GK & 55.3 & \textbf{57.7} & +2.4\% \\
MM & 29.8 & \textbf{30.1} & +0.3\% \\
AM & 37.5 & \textbf{40.0} & +2.5\% \\
CM & 37.3 & \textbf{43.2} & +5.9\% \\
MS & 72.9 & \textbf{74.1} & +1.2\% \\
MW & 96.8 & \textbf{97.5} & +0.7\% \\
OL & 28.7 & \textbf{29.3} & +0.6\% \\
AD & 95.1 & \textbf{95.2} & +0.1\% \\
MH & 69.1 & \textbf{69.2} & +0.1\% \\
\bottomrule
\end{tabular}
\caption{Performance of the PPO-based variant under the SC@21 setting.
Abbreviations: \textbf{AD}=ASDiv, \textbf{GK}=Gaokao~2023 (EN), \textbf{GS}=GSM8K,
\textbf{MH}=MATH, \textbf{MM}=Minerva\_math, \textbf{MS}=MMLU-STEM,
\textbf{CM}=College\_math, \textbf{AM}=Amc23, \textbf{MW}=MAWPS,
\textbf{OL}=OlympiadBench.}
\label{tab:ppo_variant}
\end{table}

\clearpage
\subsection{Example for Analysis}
\label{sec:appendix-c2}
Below is an exmaple for analysis.
\begin{table}[H]
\centering
\small
\begin{tabular}{@{}p{0.31\textwidth}p{0.31\textwidth}p{0.31\textwidth}@{}}
\toprule
\textbf{\methodname{}-Base} & \textbf{\methodname{}-Base} & \textbf{\methodname{}-KL} \\
\midrule
\makecell[tl]{\parbox[t]{\linewidth}{%
\textbf{Input:} Let $A$, $B$ be orthogonal matrices, $k$ be a nonzero real number, and $P$ an invertible matrix. Which of the following is true? \\
A) $A + B$ is also an orthogonal matrix \\
B) $kA$ is also an orthogonal matrix \\
C) $AB$ is also an orthogonal matrix \\
D) $P^{-1} A P$ is also an orthogonal matrix
}}
&
\makecell[tl]{\parbox[t]{\linewidth}{%
\textbf{Input:} \\
\textbf{Concepts:}\\
Orthogonal and Unitary Transformations: Orthogonal and Unitary Matrices\\
An $n$-dimensional real matrix $A$ is called an orthogonal matrix if\\
$A^{\prime}A = AA^{\prime} = I_n$.\\[0.5em]
An $n$-dimensional complex matrix $U$ is called a unitary matrix if\\
$\overline{U}^{\prime} U = U \overline{U}^{\prime} = I_n$.\\
\noindent\rule{\linewidth}{0.4pt}\\
\textbf{Exercise:}
Let $A$, $B$ be orthogonal matrices, $k$ be a nonzero real number, and $P$ an invertible matrix. Which of the following is true? \\
A) $A + B$ is also an orthogonal matrix \\
B) $kA$ is also an orthogonal matrix \\
C) $AB$ is also an orthogonal matrix \\
D) $P^{-1} A P$ is also an orthogonal matrix
}}
&
\makecell[tl]{\parbox[t]{\linewidth}{%
\textbf{Input:} Let $A$, $B$ be orthogonal matrices, $k$ be a nonzero real number, and $P$ an invertible matrix. Which of the following is true? \\
A) $A + B$ is also an orthogonal matrix \\
B) $kA$ is also an orthogonal matrix \\
C) $AB$ is also an orthogonal matrix \\
D) $P^{-1} A P$ is also an orthogonal matrix
}}
\\
 
\\
\midrule

\makecell[tl]{\parbox[t]{\linewidth}{%
\textbf{Output:} 
Answer: C \textbar{} Response: B \\
… B) $kA$ is also an orthogonal matrix. This is true. \textbf{If $A$ is an orthogonal matrix and $k$ is a nonzero real number, then $kA$ is also an orthogonal matrix. This is because multiplying an orthogonal matrix by a scalar does not change its orthogonality.} C) $AB$ is also an orthogonal matrix.This is true. If $A$ and $B$ are orthogonal matrices, then their product $AB$ is also an orthogonal matrix. This is because the product of two orthogonal matrices is orthogonal.…The final answer is boxed\{B\}. }}
&
\makecell[tl]{\parbox[t]{\linewidth}{%
\textbf{Output:} 
Answer: C \textbar{} Response: B \\
… B) $kA$ is also an orthogonal matrix This is true. \textbf{If $A$ is an orthogonal matrix, then $kA$ is also an orthogonal matrix for any nonzero real number $k$.} C) $AB$ is also an orthogonal matrix This is true. If $A$ and $B$ are orthogonal matrices, then $AB$ is also an orthogonal matrix. Therefore, the correct answer is B) $kA$ is also an orthogonal matrix.}}
&
\makecell[tl]{\parbox[t]{\linewidth}{%
\textbf{Output:} 
Answer: C \textbar{} Response: C \\
… B) $kA$ is not necessarily an orthogonal matrix. \textbf{If $k$ is a nonzero real number, then $kA$ is not orthogonal unless $k = pm 1$. This is because the columns of $kA$ are not necessarily orthogonal unit vectors.} C) $AB$ is an orthogonal matrix. The product of two orthogonal matrices is also orthogonal. This is because the columns of $AB$ are linear combinations of the columns of $A$, which are orthogonal unit vectors. Therefore, the columns of $AB$ are also orthogonal unit vectors. …The correct answer is C) $AB$ is also an orthogonal matrix. The final answer is boxed\{C\}. }}
\\
\bottomrule
\end{tabular}
\caption{This is a case study comparing the original GRPO model, a variant augmented with concept prompts, and the model after \textbf{\methodname{}-KL} training. The example illustrates that the original model tends to fall into a fixed reasoning pattern when dealing with orthogonal matrices. Even when provided with detailed conceptual derivations, the model still prefers to rely on patterns learned during pretraining. In contrast, the model trained with \textbf{\methodname{}-KL} is able to break out of this fixed paradigm and effectively apply the relevant theorems of orthogonal matrices.}
\label{tab:test_time_prompt}
\end{table}

\section{Use of LLM}
We have only used LLM for language polishing purposes in the paper writing. We do not use LLM for idea formalization, or to an extent that it could be regarded as a contributor.
\end{document}